\documentclass[reqno]{amsart}

\usepackage{graphicx}\graphicspath{{figures/}}
\usepackage{multicol}

\usepackage{caption}
\usepackage{subcaption}
\usepackage{datetime}

\usepackage[usenames]{color}

\usepackage{figures/picins}
\usepackage{tikz}
\usepackage{tikz-qtree}
\usetikzlibrary{shapes.geometric,decorations.markings,arrows,chains,matrix,positioning,scopes,calc,trees,shadows}
\tikzstyle{mybox} = [draw=white, rectangle]

\theoremstyle{plain}
\newtheorem{proposition}{Proposition}[section]

\theoremstyle{definition}
\newtheorem{definition}[proposition]{Definition}

\theoremstyle{remark}

\def\[#1\]{\begin{align}#1\end{align}}
\newcommand{\defas}{:=}
\newcommand{\given}{\mid}
\newcommand{\pars}{\,\cdot\,}

\newcommand{\Naturals}{\mathbb{N}}
\newcommand{\Rationals}{\mathbb{Q}}

\newcommand{\Reals}{\mathbb{R}}

\newcommand{\iid}{\textrm{i.i.d.}}
\newcommand{\as}{\textrm{a.s.}}

\DeclareMathOperator*{\newlim}{\mathrm{lim}\vphantom{\mathrm{infsup}}}

\DeclareMathOperator*{\newmax}{\mathrm{max}\vphantom{\mathrm{infsup}}}
\DeclareMathOperator*{\newinf}{\mathrm{inf}\vphantom{\mathrm{infsup}}}

\renewcommand{\lim}{\newlim}

\renewcommand{\max}{\newmax}
\renewcommand{\inf}{\newinf}

\newcommand{\Nats}{\mathbb{N}}

\newcommand{\st}{\,:\,}
\renewcommand{\Pr}{\mathbb{P}}
\newcommand{\defn}[1]{\emph{#1}}
\newcommand{\cD}{\mathcal{D}}

\DeclareMathOperator*{\BetaD}{Beta}
\DeclareMathOperator*{\BernoulliD}{Bernoulli}

\newcommand{\query}{\ensuremath{\mathsf{QUERY}}}

\newcommand{\newprogram}[1]{\ensuremath{\mathsf{#1}}}
\newcommand{\DS}{\newprogram{DS}}

\newcommand{\PP}{\newprogram{P}}
\newcommand{\CC}{\newprogram{C}}

\newcommand{\PA}{\textrm{Pa}}
\newcommand{\quoteskip}{[.\,.\,.]}
\newcommand{\emphasize}[1]{{{#1}}}

\usepackage{mathrsfs} 
\usepackage{outlines}

\include{myinclude}

\usepackage{amsmath, amssymb, bm}
\usepackage[square,numbers,longnamesfirst]{natbib}
\usepackage[colorlinks,citecolor=blue,urlcolor=blue,linkcolor=black]{hyperref}

\usepackage{verbatim}

\newcommand{\enode}{\bigotimes}
\newcommand{\anode}{\bigoplus}
\newcommand{\node}{\bigodot}
\DeclareMathOperator{\score}{score}

\title
[Towards common-sense reasoning via conditional simulation]
{Towards common-sense reasoning\\ via conditional simulation:\\ legacies of Turing in Artificial Intelligence}
   
  \tikzset{
  treenode/.style = {align=center, inner sep=2pt, text centered},
  ltext/.style = {treenode, rectangle, draw=none,
    minimum width=0.5em, minimum height=0.5em},
  term/.style = {treenode, rectangle, draw=none, font=\bfseries, text centered,
    minimum width=0.5em, minimum height=0.5em},
  stoc/.style = {treenode, rectangle, draw=black,
    minimum width=0.5em, minimum height=0.5em},
}

\begin{document} 
\bibliographystyle{droy-halpha} 

\author[Freer]{Cameron E.~Freer}
\address{Massachusetts Institute of Technology\\
Computer Science and Artificial Intelligence Laboratory}
\email{freer@math.mit.edu}

\author[Roy]{Daniel M.~Roy}
\address{%
University of Cambridge\\
Department of Engineering
}
\email{d.roy@eng.cam.ac.uk}

\author[Tenenbaum]{Joshua B. Tenenbaum}
\address{Massachusetts Institute of Technology\\Department of Brain and Cognitive Sciences}
\email{jbt@mit.edu}

\begin{abstract}
The problem of replicating the flexibility of human common-sense reasoning has captured the imagination of computer scientists since
the early days of Alan Turing's foundational work on computation and the philosophy of artificial intelligence.
In the intervening years, the idea of cognition as computation has emerged as a fundamental
tenet of Artificial Intelligence (AI) and cognitive science.
But what kind of computation is cognition?

We describe a computational formalism centered around a probabilistic Turing machine called \query, which captures the operation of probabilistic conditioning via \emph{conditional simulation}.
Through several examples and analyses, we demonstrate how the 
\query\ abstraction can be used to cast common-sense reasoning as
probabilistic inference in a statistical model of our observations and the uncertain
structure of the world that generated that experience.
This formulation is a recent synthesis of several research programs in AI and cognitive science, but it also
represents a surprising convergence of several of Turing's
pioneering insights in AI, the foundations of computation, and statistics.
\end{abstract}

\maketitle

\renewcommand\contentsname{\!\!\!\!}
\setcounter{tocdepth}{1}
\vspace{-8mm}\tableofcontents
 
\section{Introduction}

In his landmark paper \emph{Computing Machinery and Intelligence}
\cite{Tur50}, Alan Turing predicted
that by the end of the twentieth century, 
``general educated opinion will have altered so much that one will be able to speak of machines thinking without expecting to be contradicted.''  Even if Turing has not yet been proven right,
 the idea of \emph{cognition as computation} has emerged as
a fundamental tenet of Artificial Intelligence (AI) and cognitive science. 
But what kind of computation---what kind of computer program---is cognition?

AI researchers have made impressive progress since the birth of the field over 60 years ago. 
Yet despite this progress, no existing AI system can reproduce any nontrivial fraction of the inferences made regularly by children.
Turing himself appreciated that matching the capability of children, 
e.g., in language,
presented a key challenge for AI:
\begin{quote}
We hope that machines will eventually compete with men in all purely intellectual fields.  
But which are the best ones to start with?  
Even this is a difficult decision.
Many people think that a very abstract activity, like the playing of chess, would be best. 
It can also be maintained that it is best to 
provide the machine with the best sense organs money can buy, and then
teach it to understand and speak English. This process could follow the
normal teaching of a child. Things would be pointed out and named, etc. Again I do not know what the right
answer is, but I think both approaches should be tried.
\cite[p.~460]{Tur50}
\end{quote}
Indeed, many of the problems once considered to be grand AI challenges have fallen prey to essentially brute-force algorithms backed by enormous amounts of computation, often robbing us of the insight we hoped to gain by studying these challenges in the first place.  Turing's presentation of his ``imitation game'' (what we now call ``the Turing test''), and the problem of common-sense reasoning implicit in it, demonstrates that he understood the difficulty inherent in the open-ended, if commonplace, tasks involved in conversation.  Over a half century later, the Turing test remains resistant to attack.

The analogy between minds and computers has spurred incredible scientific progress in both directions, but there are still fundamental disagreements about the nature of the computation performed by our minds, and how best to narrow the divide between the capability and flexibility of human and artificial intelligence.  The goal of this article is to describe a computational formalism that has proved useful for building simplified models of common-sense reasoning.  The centerpiece of the formalism is a universal probabilistic Turing machine called \query\ that performs \emph{conditional simulation}, and thereby captures the operation of conditioning probability distributions that are themselves represented by probabilistic Turing machines.  We will use \query\ to model the inductive leaps that typify common-sense reasoning.  The distributions on which \query\ will operate are models of latent unobserved processes in the world and the sensory experience and observations they generate.  Through a running example of medical diagnosis, we aim to illustrate the flexibility and potential of this approach.

The \query\ abstraction is a component of several research programs in AI and cognitive science developed jointly with a number of collaborators.  This chapter represents our own view on a subset of these threads and their relationship with Turing's legacy.  Our presentation here draws heavily on both the work of \emphasize{Vikash Mansinghka} on ``natively probabilistic computing''
\cite{Man09, MJT08, Man11, MR13}
and the ``probabilistic language of thought'' hypothesis proposed and developed by \emphasize{Noah Goodman}
\cite{KGT08,GTFG08,GG12, GT12}.
Their ideas form core aspects of the picture we present.  The \emph{Church}
probabilistic programming language (introduced in
\cite{GMRBT08} by Goodman, Mansinghka, Roy, Bonawitz, and
Tenenbaum) and various Church-based cognitive science tutorials (in
particular, \cite{GTO11}, developed by Goodman, Tenenbaum, and O'Donnell) have also had a strong influence on the presentation.

This approach also draws from work in cognitive science on ``theory-based \linebreak Bayesian models'' of inductive learning and reasoning
\cite{TGK06}
due to Tenenbaum and various collaborators
\cite{GKT08, KT08, TKGG11}.
Finally, some of the theoretical aspects that we present are based on results in computable probability theory by Ackerman, Freer, and Roy
\cite{Roy11, AFR11}.

While the particular formulation of these ideas is recent, they have antecedents in much earlier work on
the foundations of computation and computable analysis, common-sense reasoning in AI, and Bayesian modeling
and statistics. In all of these areas, Turing had pioneering insights.

\subsection{A convergence of Turing's ideas}\label{sec:converge}

In addition to Turing's well-known contributions to the philosophy of AI,
many other aspects of his work---across statistics, the foundations of computation, and even morphogenesis---have converged in the modern study of AI.
In this section, we highlight a few key ideas that will frequently surface
during our account of common-sense reasoning via conditional simulation.

An obvious starting point is Turing's own proposal for a research program to pass his eponymous test.
From a modern perspective, Turing's focus on learning (and in particular, induction) was especially
prescient.  For Turing, the idea of programming an intelligent machine entirely by hand was clearly
infeasible, and so he reasoned that it would be necessary to construct a machine with the ability to adapt its own behavior in light of experience---i.e., with the ability to learn:
\begin{quote}
Instead of trying to produce a programme to simulate the adult mind, why not rather try to produce one that simulates the child's?  If this were then subjected to an appropriate course of education one would obtain the adult brain.  \cite[p.~456]{Tur50}
\end{quote}
Turing's notion of learning was inductive as well as deductive, in contrast to much of the work that followed in the first decade of AI.
In particular, he was quick to explain that such a machine would have its flaws (in reasoning, quite apart
from calculational errors):
\begin{quote}
[A machine] might have some method for drawing conclusions by scientific induction. We must expect such a method to lead occasionally to erroneous results.
\cite[p.~449]{Tur50}
\end{quote}
Turing also appreciated that a machine would not only have to learn facts, but would also need to learn \emph{how to learn}:
\begin{quote}
Important amongst such imperatives will be ones which regulate the order in which the rules of the logical system concerned are to be applied. 
For at each stage when one is using a logical system, there is a very large number of alternative steps, any
of which one is permitted to apply
\quoteskip\ 
These choices make the difference between a brilliant and a footling reasoner, not the difference between a sound and a fallacious one. 
\quoteskip\
[Some such imperatives]
may be `given by authority', but others may be produced by the machine itself,  \emph{e.g.} by  scientific induction.
\cite[p.~458]{Tur50}
\end{quote}
In addition to making these more abstract points, Turing
presented
a number of concrete proposals for how a machine might be programmed to learn.  
His ideas capture the essence of supervised, unsupervised, and reinforcement learning, each major areas in modern AI.\footnote{Turing also developed some of the early ideas regarding \emph{neural networks}; see the discussions in \cite{Tur48} about ``unorganized machines'' and their education and organization. This work, too, has grown into a large field of modern research, though we will not explore neural nets in the present article. For more details, and in particular the connection to work of McCulloch and Pitts \cite{MP43}, see Copeland and Proudfoot \cite{CP96} and Teuscher \cite{Teu02}.}
In Sections~\ref{sec:latent} and \ref{sec:tom} we will return to Turing's writings on these matters.

One major area of Turing's contributions, while often overlooked, is statistics.  
In fact, Turing, along with I.~J.\ Good, made key advances in statistics 
in the course of breaking the Enigma during World War~II.
Turing and Good
developed
new techniques for incorporating evidence and new approximations for
estimating parameters in hierarchical models \cite{Goo79, Goo00} 
(see also \cite[\S5]{Zab95} and \cite{Zab12}),
which
were 
among
the most important applications  
of Bayesian statistics at the time \cite[\S3.2]{Zab12}.
Given Turing's interest in learning machines and his deep understanding of statistical methods, 
it would have been intriguing to see a proposal to combine the two areas.  
Yet if he did consider these connections, it seems he never published such work.
On the other hand, much of modern AI rests upon a statistical
foundation, including Bayesian methods. This perspective permeates the approach we will describe, wherein learning is achieved via Bayesian inference,
and in Sections~\ref{sec:latent} and \ref{sec:definetti} we will re-examine some of Turing's wartime statistical work
in the context of hierarchical models.

A core latent
hypothesis underlying Turing's diverse body of work 
was that processes in nature, including our minds, could be understood through 
mechanical---in fact, \emph{computational}---descriptions.
One of Turing's crowning achievements was his introduction of the $a$-machine, which we now call the Turing
machine.  The Turing machine characterized the limitations and possibilities of computation by providing
a mechanical description of a human computer. 
Turing's work on
morphogenesis \cite{Tur52} and AI each sought mechanical explanations in still further domains.
Indeed, in all of these areas, Turing was acting as a natural
scientist \cite{Hod97}, building models of natural phenomena
using the language of computational processes.

In our account of common-sense reasoning as conditional simulation,
we will use probabilistic Turing machines to 
represent
mechanical descriptions of the world,
much like those Turing sought.
In each case, the stochastic machine 
represents one's
uncertainty about the generative process giving rise to some pattern in the natural world.
This description then enables probabilistic inference (via \query) about these patterns, 
allowing us to make decisions and manage our uncertainty in light of new evidence.
Over the course of the article we will see a number of 
stochastic generative processes 
of increasing sophistication, culminating in models of decision making that rely crucially on recursion.
Through its emphasis on inductive learning, Bayesian statistical techniques, universal computers,
and mechanical models of nature,
this approach to common-sense reasoning 
represents a convergence of many of Turing's ideas.

\subsection{Common-sense reasoning via \query}

For the remainder of the paper, our focal point will be the probabilistic Turing machine  \query, which
implements a generic form of probabilistic conditioning.
\query\ allows one to make predictions using
complex probabilistic models that are themselves 
specified using
probabilistic Turing machines.  
By using \query\ appropriately, one can describe various forms of learning, inference, and 
decision-making. These arise via Bayesian inference, and common-sense behavior can be seen to follow
\emph{implicitly} from past experience and models of causal structure and goals, rather than explicitly via rules or purely deductive reasoning.
Using the extended example of medical diagnosis,
we 
aim to demonstrate that \query\ is a surprisingly powerful abstraction for expressing common-sense reasoning tasks that have, until recently, largely defied formalization.

As with Turing's investigations in AI, the approach we describe has been motivated by reflections on
the details of human cognition, as well as on the nature of computation.
In particular, much of the AI framework
that we describe has been inspired by research in cognitive science attempting to 
model human inference and learning. 
Indeed, hypotheses expressed in this framework have been compared 
with the judgements
and behaviors
of
human children and
adults
in many psychology experiments.
Bayesian generative models,
of the sort we describe here, have been shown to 
predict human behavior on a
wide range of cognitive tasks, often with high quantitative accuracy.
For examples of such models and the corresponding experiments, see the review article
\cite{TKGG11}. We will return to some of these more complex models in Section~\ref{conclusion}.
We now proceed to define \query\ and illustrate its use via increasingly complex problems and the questions
these raise.

\section{Probabilistic reasoning and \query}
\label{sec:DS}

The specification of a probabilistic model can implicitly define a space of complex and useful
behavior.
In this section we informally describe the universal probabilistic
Turing machine \query, and then use \query\ to explore a medical diagnosis example that captures many aspects of common-sense reasoning, but in a simple domain.  Using \query, we highlight the role of conditional independence and conditional probability in building compact 
yet highly flexible systems.

\subsection{An informal introduction to \query}

The \query\ formalism 
was originally developed in the context of the Church probabilistic programming language \linebreak
\cite{GMRBT08}, and has been further explored by Mansinghka \cite{Man11} and
Mansinghka and Roy \cite{MR13}.

At the heart of the \query\ abstraction is a probabilistic variation of
Turing's own mechanization \cite{Tur36}
of the capabilities of human ``computers'', the Turing machine.
A Turing machine is a finite automaton with read, write, and seek access to a finite collection of infinite binary tapes, which it may use throughout the course of its execution. 
Its input is loaded onto one or more of its tapes prior to execution, and the output 
is the content of (one or more of) its tapes after the machine enters its halting state.
A \emph{probabilistic} Turing machine (PTM) is simply a Turing machine with an additional read-only tape comprised of a sequence of independent random bits, which the finite automaton may read and use as a source of randomness.

\newcommand{\UNIVERSAL}{\newprogram{UNIVERSAL}}
Turing machines (and their probabilistic generalizations) capture a robust notion of deterministic (and probabilistic) computation:  
Our use of the Turing machine abstraction relies on the remarkable existence of \emph{universal} Turing machines, which can simulate all other Turing machines.
 More precisely,  there is a PTM \UNIVERSAL\ and an encoding $\{e_s \st s \in \{0,1\}^*\}$ 
of all PTMs, 
where $\{0,1\}^*$ denotes the set of finite binary strings,
such that, on inputs $s$ and $x$, \UNIVERSAL\ halts and outputs the string $t$ if and only if (the Turing machine encoded by) $e_s$ halts and outputs $t$ on input $x$.
Informally, the input $s$ to \UNIVERSAL\ is analogous to a program written in a programming language, and so we will speak of (encodings of) Turing machines and programs interchangeably.
 
\query\ is a PTM that takes two inputs, called the \emph{prior program} $\PP$ and \emph{conditioning predicate} $\CC$, both of which are themselves (encodings of) PTMs that take no input (besides the random bit tape), with the further restriction that the predicate $\CC$ return only a $1$ or $0$ as output.   The semantics of \query\ are straightforward:
first generate a sample from $\PP$; if $\CC$ is satisfied,
then output the sample; otherwise, try again. More precisely:
 \begin{enumerate}
 \item Simulate the predicate $\CC$ on a random bit tape $R$ (i.e., using the existence of a universal
Turing machine, 
determine the output of the PTM $\CC$, if $R$ were its random bit tape); 
 \item If (the simulation of) $\CC$ produces $1$ (i.e., if $\CC$ \emph{accepts}), then simulate and return the output produced by $\PP$, using the \emph{same} random bit tape $R$;
 \item Otherwise (if $\CC$ \emph{rejects} $R$, returning 0), return to step 1, using 
an independent 
sequence $R'$ of random bits.
 \end{enumerate}
It is important to stress that $\PP$ and $\CC$ share a random bit tape on each iteration, and so the predicate $\CC$ may, in effect, act as though it has access to any intermediate value computed by the prior program $\PP$ when deciding whether to accept or reject a random bit tape.  More generally, any value computed by $\PP$ can be recomputed by $\CC$ and vice versa.
We will use this fact to simplify the description of predicates, informally
referring to values computed by $\PP$ in the course of defining a
\linebreak predicate $\CC$.

As a first step towards understanding \query,  note that if $\top$ is a PTM that always accepts (i.e., always outputs $1$), then $\query(\PP,\top)$ produces the same distribution on outputs as executing $\PP$ itself, as the semantics imply that $\query$ would halt on the first iteration. 

\newcommand{\NNN}{\newprogram{N_{180}}}
\newcommand{\DIV}{\newprogram{DIV_{2,3,5}}}
Predicates that are not identically $1$ lead to more interesting behavior.
Consider the following simple example based on a remark by Turing \cite[p.~459]{Tur50}:
Let $\NNN$ be a PTM that returns (a binary encoding of) an integer $N$ drawn uniformly at random in the range 1 to 180, and let $\DIV$ be a PTM that accepts (outputs 1) if $N$ is divisible by 2, 3, and 5; and rejects (outputs 0) otherwise.
Consider a typical output of
\begin{align*}
\query(\NNN,\DIV).
\end{align*}
Given the semantics of \query, we know that the output will fall in the set 
\begin{align}\label{simpleset}\{30, 60, 90, 120, 150,180\}\end{align} 
and moreover, 
because each of these possible values of $N$ was \emph{a priori} equally likely to arise from executing $\NNN$ alone, this remains true \emph{a posteriori}.
You may recognize this as the conditional distribution of a uniform distribution conditioned to lie in the
set \eqref{simpleset}.  
Indeed, \query\ performs the operation of conditioning a distribution.

The behavior of \query\ can be described more formally with 
notions from probability theory.
In particular, from this point on, we will think of the output of a PTM (say, $\PP$) as a random variable (denoted by $\varphi_{\PP}$) defined on an underlying probability space with probability measure $\Pr$.  
(We will define this probability space formally in Section~\ref{sec:formal}, but informally it represents the random bit tape.)   When it is clear from context, we will also regard any named intermediate value (like $N$) as a random variable on this same space.  
Although Turing machines manipulate binary representations, we will often gloss over the details of how elements of other countable sets (like the integers, naturals, rationals, etc.) are represented in binary, but only when there is no risk of serious misunderstanding.

In the context of $\query(\PP,\CC)$, the output distribution of $\PP$, which can be written $\Pr(\varphi_{\PP} \in \cdot\,)$, is called the \defn{prior} distribution.  
Recall that, for all measurable sets (or simply \emph{events}) $A$ and $C$ , 
\begin{align}\label{simplecond}
\Pr( A \given C) 
\defas \frac{\Pr( A \cap C)}{\Pr(C)},
\end{align}
is the conditional probability of the event $A$ given the event $C$, provided that $\Pr(C) > 0$.
Then the distribution of the output of $\query(\PP,\CC)$, called the \defn{posterior} 
distribution of $\varphi_{\PP}$, is the conditional distribution of $\varphi_{\PP}$ given the event $\varphi_{\CC} = 1$, written 
\begin{align*}\Pr(\varphi_{\PP} \in \cdot \given \varphi_{\CC} = 1).  \end{align*} 

Then returning to our example above,
the prior distribution, $\Pr(N \in \cdot\,)$, is the uniform distribution on the set $\{1,\dotsc,180\}$, and the posterior distribution, 
\begin{align*}\Pr(N \in \cdot \given \text{$N$ divisible by 2, 3, and 5}),\end{align*} is the uniform distribution on the set given in \eqref{simpleset}, as can be verified via equation~\eqref{simplecond}.

Those familiar with statistical algorithms will recognize the mechanism of \query\ to be exactly that of a so-called ``rejection sampler''.  
Although the definition of \query\ describes an explicit algorithm,
we do not 
actually
intend \query\ to be executed in practice, but rather intend for it to 
define and represent complex distributions.  (Indeed, the description can be used by algorithms that work by very different methods than rejection sampling, and can aid in the communication of ideas between researchers.)

The actual implementation of \query\ in more
efficient ways than via a rejection sampler is an active area of research, especially via techniques involving
Markov chain Monte Carlo (MCMC); see, e.g., \cite{GMRBT08, WSG11, WGSS11, SG12}. 
Turing himself recognized the potential usefulness of randomness in computation, suggesting:
\begin{quote}
It is probably wise to include a random element in a learning machine.  
A random element is rather useful when we are searching for a solution of some problem.
\cite[p.~458]{Tur50}
\end{quote}
Indeed, some aspects of these algorithms are strikingly reminiscent of Turing's description of a random system of
rewards and punishments in guiding the organization of a machine:
\begin{quote}
The character may be subject to
some random variation. Pleasure interference has a tendency to fix the character
i.e.\ towards preventing it changing, whereas pain stimuli tend to disrupt the
character, causing features which had become fixed to change, or to become
again subject to random variation.
\cite[\S10]{Tur48}
\end{quote}
However, in this paper, we will not go into further details of implementation, nor the host of interesting computational
questions this endeavor raises.

Given the subtleties of conditional probability, it will often be helpful to keep in mind the behavior of a rejection-sampler when considering examples
of \query.
(See \cite{SG92} for more examples of this approach.)
Note that, in our example, every simulation of $\NNN$ generates a number ``accepted by'' $\DIV$
with probability $\frac 1 {30}$, and so, on average, we would expect the loop within \query\ to repeat approximately $30$ times before halting. 
However, there is no finite bound on how long the computation could run.
On the other hand, one can show that $\query(\NNN,\DIV)$ eventually halts with probability one (equivalently, it halts \emph{almost surely}, sometimes abbreviated ``a.s.'').

Despite the apparent simplicity of the \query\ construct, we will see that it captures the essential structure of a range of common-sense inferences.  
We now demonstrate the power of the \query\ formalism by exploring its behavior in a medical diagnosis example.

\subsection{Diseases and their symptoms}
Consider the following prior program, \DS, \linebreak which represents a simplified model of the pattern of \emph{Diseases and Symptoms} we might find in a typical patient chosen at random from the population.  
At a high level, the model posits that the patient may be suffering from some, possibly empty, set of diseases, and that these diseases can cause symptoms.  The prior program \DS\ proceeds as follows:
For each disease $n$ listed in Table~\ref{tab:diseases}, sample an independent binary random variable $D_n$ with mean $p_n$, which we will interpret as indicating whether or not a patient has disease $n$ depending on whether $D_n=1$ or $D_n=0$, respectively.
\begin{table}[t]
\hrule
\vspace{7pt}
\begin{tiny}
\begin{subtable}[t]{0.27 \textwidth}
\subcaption{Disease marginals}
\begin{tabular}{l|c|l}
$n$&Disease&$p_n$\\ \hline
1& Arthritis&0.06\\
2& Asthma&0.04\\
3& Diabetes&0.11\\
4& Epilepsy&0.002\\
5& Giardiasis&0.006\\
6& Influenza&0.08\\
7& Measles& 0.001\\
8& Meningitis&0.003\\
9& MRSA&0.001\\
10& Salmonella&0.002\\
11& Tuberculosis&0.003
\end{tabular}
\label{tab:diseases}
\end{subtable}
\begin{subtable}[t]{0.35 \textwidth}
\subcaption{Unexplained symptoms}
\centering
\begin{tabular}{l|c|l}
$m$&Symptom&$\ell_m$\\ \hline
1& Fever&0.06\\
2& Cough&0.04\\
3& Hard breathing&0.001\\
4& Insulin resistant&0.15\\
5& Seizures&0.002\\
6& Aches&0.2\\
7& Sore neck&0.006
\end{tabular}
\label{tab:symptoms}
\end{subtable}
\begin{subtable}[t]{0.35 \textwidth}
\subcaption{Disease-symptom rates}
\begin{tabular}{r|ccccccc}
$ c_{n,m}$ &1&2&3&4&5&6&7\\\hline
1&.1&.2&.1&.2&.2&.5&.5\\ 
2&.1&.4&.8&.3&.1&.0&.1\\ 
3&.1&.2&.1&.9&.2&.3&.5\\ 
4&.4&.1&.0&.2&.9&.0&.0\\ 
5&.6&.3&.2&.1&.2&.8&.5\\ 
6&.4&.2&.0&.2&.0&.7&.4\\ 
7&.5&.2&.1&.2&.1&.6&.5\\ 
8&.8&.3&.0&.3&.1&.8&.9\\ 
9&.3&.2&.1&.2&.0&.3&.5\\ 
10&.4&.1&.0&.2&.1&.1&.2\\ 
11&.3&.2&.1&.2&.2&.3&.5
\end{tabular}
\label{tab:rates}
\end{subtable}
\end{tiny}
\vspace{2mm}
\caption{Medical diagnosis parameters. (These values are fabricated.)  (a) $p_n$ is the marginal probability that a patient has a disease $n$.  (b) $\ell_m$ is the probability that a patient presents symptom $m$, assuming they have no disease. (c) $c_{n,m}$ is the probability that disease $n$ causes  symptom $m$ to present, assuming the patient has disease $n$.}
\label{tab:probs}
\vspace{-1.5mm}
\hrule
\end{table}
For each symptom $m$ listed in Table~\ref{tab:symptoms}, sample an independent binary random variable $L_m$ with mean $\ell_m$
and for each pair $(n,m)$ of a disease and symptom, sample an independent binary random variable $C_{n,m}$
with mean $c_{n,m}$, as listed in Table~\ref{tab:rates}. 
(Note that the numbers in all three tables have been fabricated.)
Then, for each symptom $m$, define 
\begin{align*}
S_m = \max \{ L_m, D_1\cdot C_{1,m},\dotsc, D_{11} \cdot C_{11,m} \},
\end{align*}
so that $S_m \in \{0,1\}$.  We will interpret $S_m$ as indicating that a patient has symptom $m$; 
the definition of $S_m$ implies that this 
holds
when any of the variables on the right hand side take the value $1$.  (In other words, the $\max$ operator is playing the role of a logical OR operation.)  
Every term of the form $D_n \cdot C_{n,m}$ is interpreted as indicating whether (or not) the patient has disease $n$ \emph{and} disease $n$ has caused symptom $m$.  
The term $L_m$ captures the possibility that the symptom may present itself despite the patient having none of the listed diseases.
Finally,
define the output of \DS\ to be the vector $(D_1,\dotsc,D_{11},S_1,\dotsc,S_7)$.

If we execute \DS, or equivalently $\query(\DS,\top)$, then we might see outputs like those in the following
array:
\begin{small}
\begin{center}
\begin{tabular}{l|ccccccccccc|ccccccc}
\multicolumn{1}{c}{}&
\multicolumn{11}{c}{Diseases}&
\multicolumn{7}{c}{Symptoms}\\
&
$\!_{1}$&
$\!_{2}$&
$\!_{3}$&
$\!_{4}$&
$\!_{5}$&
$\!_{6}$&
$\!_{7}$&
$\!_{8}$&
$\!_{9}$&
$\!_{{10}}$&
$\!_{{11}}$&
$\!_{1}$&
$\!_{2}$&
$\!_{3}$&
$\!_{4}$&
$\!_{5}$&
$\!_{6}$&
$\!_{7}$\\\hline
$\,^{\!_{1}}$&
0&0&0&0&0&0&0&0&0&0&0&  0&0&0&0&0&0&0 \\
$\,^{\!_{2}}$&
0&0&0&0&0&0&0&0&0&0&0&  0&0&0&0&0&0&0 \\
$\,^{\!_{3}}$&
0&0&1&0&0&0&0&0&0&0&0&  0&0&0&1&0&0&0 \\
$\,^{\!_{4}}$&
0&0&1&0&0&1&0&0&0&0&0&  1&0&0&1&0&0&0 \\
$\,^{\!_{5}}$&
0&0&0&0&0&0&0&0&0&0&0&  0&0&0&0&0&1&0 \\
$\,^{\!_{6}}$&
0&0&0&0&0&0&0&0&0&0&0&  0&0&0&0&0&0&0 \\
$\,^{\!_{7}}$&
0&0&1&0&0&0&0&0&0&0&0&  0&0&0&1&0&1&0 \\
$\,^{\!_{8}}$&
0&0&0&0&0&0&0&0&0&0&0&  0&0&0&0&0&0&0 \\
\end{tabular}
\end{center}
\end{small}
We will interpret the rows as representing eight patients chosen independently at random, 
the first two free from disease and not presenting any symptoms; 
the third suffering from diabetes and presenting insulin resistance; 
the fourth suffering from diabetes and influenza, and presenting a fever and insulin resistance; 
the fifth suffering from unexplained aches; 
the sixth free from disease and symptoms; 
the seventh suffering from diabetes, and presenting  insulin resistance and aches; 
and the eighth also disease and symptom free.

This model is a toy version of the real diagnostic model
QMR-DT \cite{SMH+91}. QMR-DT is probabilistic model with essentially the structure of \DS,
built from data in the Quick
Medical Reference (QMR) knowledge base of hundreds of diseases and
thousands of findings (such as symptoms or test results).
A key aspect of this model is the disjunctive relationship between the diseases and the symptoms, known as a
``noisy-OR'', which remains a popular modeling idiom. 
In fact, the structure of this model, and in particular the idea of layers
of disjunctive causes, goes back even further to the ``causal calculus''
developed by Good \cite{Goo61}, which was based in part on his wartime work
with Turing on the weight of evidence, as discussed by Pearl \cite[\S70.2]{Pea04}.

Of course, as a model of the latent processes explaining natural patterns of diseases and symptoms in a random patient, 
\DS\ still leaves much to be desired.  For example, the model assumes that
the presence or absence of any two diseases is independent, although, 
as we will see later on in our analysis, diseases are (as expected) typically not 
independent conditioned on symptoms.
On the other hand, an actual disease might cause another disease, or
might cause a symptom that itself causes another disease, possibilities
that this model does not capture.
Like QMR-DT, the model \DS\ avoids simplifications made by many earlier expert systems and probabilistic
models to not allow for the simultaneous occurrence of multiple diseases \cite{SMH+91}. 
These caveats notwithstanding, a close inspection of this simplified model will demonstrate a surprising range of common-sense reasoning phenomena.

\newcommand{\OS}{\newprogram{OS}}
Consider a predicate $\OS$, for \emph{Observed Symptoms}, that accepts if and only if $S_1=1$ and $S_7=1$, i.e., if and only if the patient presents the symptoms of a fever and a sore neck.  
What outputs should we expect from $\query(\DS,\OS)$?  
Informally, if we let $\mu$ denote the distribution 
over the combined outputs of \DS\ and \OS\ on a shared random bit tape, and
let $A = \{ (x,c) \st c = 1 \}$ denote the set of those pairs that \OS\ accepts,
then $\query(\DS,\OS)$ generates samples from the conditioned distribution $\mu(\cdot \given A)$.
Therefore, to see
what the condition $S_1 = S_7 = 1$ implies about the plausible execution
of \DS, we must consider the conditional distributions of the diseases given the symptoms.
The following conditional probability calculations may be very familiar to some readers, but will be less so to others, and so we present them here to give a more complete picture of the behavior of \query.

\subsubsection{Conditional execution}
Consider a $\{0,1\}$-assignment $d_n$ for each disease $n$,
and write $D = d$ to denote the event that $D_n = d_n$ for every such $n$.
Assume for the moment that $D=d$.
Then what is the probability that $\OS$ accepts?
The probability we are seeking is the conditional probability
\begin{align}
\Pr(S_1 = S_7 = 1 \given D = d) 
\ = \ \Pr(S_1 = 1 \given D = d)
\cdot \Pr(S_7 = 1 \given D = d), \label{Cprob}
\end{align}
where the equality follows from the observation that once the $D_n$ variables are fixed, the variables $S_1$ and $S_7$ are independent.  
Note that $S_m = 1$ if and only if $L_m = 1$ or $C_{n,m} = 1$ for some $n$ such that $d_n = 1$. 
(Equivalently, $S_m = 0$ if and only if $L_m = 0$ and $C_{n,m}=0$ for all $n$ such that $d_n = 1$.)  By the independence of each of these variables, it follows that
\begin{align}\label{scondd}
\Pr (S_m = 1 | D = d) = 1 - (1-\ell_m) \prod_{n \st d_n = 1} (1-c_{n,m}).
\end{align}
Let $d'$ be an alternative $\{0,1\}$-assignment.  We 
can now
characterize the \emph{a posteriori} odds 
\begin{align*}
\frac
{ \Pr(D = d \given S_1 = S_7 = 1) }
{ \Pr(D = d' \given S_1 = S_7 = 1) }
\end{align*}
of the assignment $d$ versus the assignment $d'$.  By Bayes' rule,
this can be rewritten as
\begin{align}\label{odds}
\frac
{\Pr(S_1 = S_7 = 1 \given D = d) \cdot \Pr (D = d)}
{\Pr(S_1 = S_7 = 1 \given D = d') \cdot \Pr (D = d')},
\end{align}
where $\Pr (D = d) = \prod_{n=1}^{11} \Pr (D_n = d_n)$ by independence.
Using \eqref{Cprob}, \eqref{scondd} and \eqref{odds}, one may calculate that
\begin{align*}
\frac
{\Pr(\textrm{Patient only has influenza} \given S_1 = S_7 = 1)}
{\Pr(\textrm{Patient has no listed disease} \given S_1 = S_7 = 1)}
\approx 42, 
\end{align*}
i.e., it is forty-two times more likely that an execution of \DS\ satisfies the predicate $\OS$ via an execution that posits the patient only has the flu than an execution which posits that the patient has no disease at all.  On the other hand, 
\begin{align*}
\frac
{\Pr(\textrm{Patient only has meningitis} \given S_1 = S_7 = 1)}
{\Pr(\textrm{Patient has no listed disease} \given S_1 = S_7 = 1)}
\approx 6, 
\end{align*}
and so
\begin{align*}
\frac
{\Pr(\textrm{Patient only has influenza} \given S_1 = S_7 = 1)}
{\Pr(\textrm{Patient only has meningitis} \given S_1 = S_7 = 1)}
\approx 7, 
\end{align*}
and hence we would expect, over many executions of $\query(\DS,\OS)$, to see roughly seven 
times as many explanations positing only influenza than positing only meningitis.

Further investigation reveals some subtle aspects of the model.  For example, consider the fact that
\begin{align}\label{explaining}
&\frac
{\Pr(\textrm{Patient only has meningitis and influenza} \given S_1 = S_7 = 1)}
{\Pr(\textrm{Patient has meningitis, maybe influenza, but nothing else} \given S_1 = S_7 = 1)}
\nonumber
\\\quad &= 0.09 \approx \Pr(\textrm{Patient has influenza}),
\end{align}
which demonstrates that, once we have observed some symptoms, diseases are no longer independent. Moreover,
this shows that once the symptoms have been ``explained'' by meningitis, there is little pressure to posit further causes, and so the posterior probability of influenza is nearly the prior probability of influenza.  This phenomenon is well-known and is called \emph{explaining away}; it is also known to be linked to the computational hardness of computing probabilities (and generating samples as \query\ does) in models of this variety.
For more details, see \cite[\S2.2.4]{Pea88}.

\subsubsection{Predicates give rise to diagnostic rules}

These various 
conditional probability calculations, and their ensuing explanations,
all follow from an analysis of the $\DS$ model conditioned on one particular (and rather simple) predicate $\OS$. Already, this 
gives rise to a picture of how $\query(\DS, \OS)$ implicitly captures an elaborate system of rules for what to believe
following the observation of a fever and sore neck in a patient, assuming the background knowledge captured in the $\DS$ program and its parameters.
In a similar way, 
every diagnostic scenario (encodable as a
predicate) gives rise to its own complex set of inferences, each expressible using \query\ and the
model $\DS$. 

As another example, if we look (or test) for the remaining symptoms and find them to all be absent, our new beliefs are captured by $\query(\DS,\OS^\star)$ where the predicate $\OS^\star$ accepts if and only if  $(S_1=S_7=1) \wedge (S_{2} = \dotsm = S_{6}=0)$.

We need not limit ourselves to reasoning about diseases given symptoms.  Imagine that we perform a diagnostic test that rules out meningitis.  We could represent our new knowledge using a predicate capturing the condition
\begin{align*}
(D_8=0) \wedge (S_1=S_7=1) \wedge (S_{2} = \dotsm = S_{6}=0).
\end{align*}
Of course this approach would not take into consideration
our uncertainty regarding the accuracy or mechanism of the diagnostic test itself, and so, ideally, we might expand the $\DS$ model to account for how the outcomes of diagnostic tests are affected by
the presence of other diseases or symptoms.  In Section~\ref{sec:definetti}, we will discuss how such an extended model might be learned from data, rather than constructed by hand.

We can also reason in the other direction, about symptoms given diseases. 
For example, public health officials might wish to know about how frequently those with influenza present no symptoms. This is captured by the conditional probability
\begin{align*}
\Pr(S_{1} = \dotsm = S_{7} = 0 \given D_6 = 1),
\end{align*}
and, via \query, by the predicate for the condition $D_6 = 1$.
Unlike the earlier examples where we reasoned backwards from effects (symptoms) to their likely causes (diseases), here we are reasoning in the same forward direction as the model \DS\ is expressed.

The possibilities are effectively inexhaustible, including more complex states of knowledge such as, \emph{there are at least two symptoms present}, or \emph{the patient does not have both salmonella and tuberculosis}.  
In Section~\ref{sec:condindep} we will consider the vast number of predicates and the resulting inferences supported by \query\ and \DS, and contrast this with the compact size of $\DS$ and the table of parameters.

In this section, we have illustrated the basic behavior of \query, and have begun to explore how it can be used to decide what to believe in a given scenario. 
These examples also demonstrate that rules governing behavior 
need not be explicitly described as rules, but can arise
implicitly via other mechanisms, like \query, paired with an appropriate prior and predicate. 
In this example, the diagnostic rules were determined by the definition of \DS\ and the table of its parameters. In Section~\ref{sec:latent}, we will examine how such a table of probabilities itself might be learned.
In fact, even if the parameters are learned from data, the structure of \DS\ itself still posits a strong structural relationship among the
diseases and symptoms. In
Section~\ref{sec:definetti} we will explore how this structure could be learned.
Finally, many common-sense reasoning tasks involve making a \emph{decision}, and not just determining what to believe.
In Section~\ref{sec:tom}, we will describe how to use \query\ to make decisions under uncertainty.

Before turning our attention to these more complex uses of \query, we pause to consider
a number of interesting theoretical questions:
What kind of probability distributions can be represented by PTMs that generate samples?  What kind of conditional distributions can be represented by \query? Or represented by PTMs in general?
In the next section we will see how Turing's work formed the foundation of the study of these questions many decades later.

\section{Computable probability theory}

We now examine the \query\ formalism in more detail, by introducing aspects of the framework of
computable probability theory, which provides rigorous notions of computability for probability
distributions, as well as the tools necessary to identify
probabilistic operations that can and cannot be performed by algorithms.
After giving a
formal description of probabilistic Turing machines and \query, we relate
them to the concept of a computable measure on a countable space.  
We then explore the representation of points (and random points) in
uncountable spaces, and examine how to use \query\ to define models over uncountable spaces like the reals.
Such models are commonplace in statistical practice, and thus might be expected to be useful for building a statistical mind.  In fact, no generic and computable \query\ formalism exists for conditioning on observations taking values in
uncountable spaces, but there are certain 
circumstances in which we can perform probabilistic inference in uncountable spaces.  

Note that although the approach we describe uses a universal Turing
machine \linebreak (\query), which can take an arbitrary pair of programs as its prior and
predicate, we do not make use of a so-called \emph{universal} prior program (itself
necessarily noncomputable). For a survey of approaches to inductive
reasoning involving a
universal prior, such as Solomonoff induction \cite{Sol64},
and computable approximations thereof, see Rathmanner and Hutter \cite{RH11}.

Before we discuss the capabilities and limitations of \query, 
we give a formal definition of \query\ in terms
of probabilistic Turing machines and conditional distributions.

\subsection{A formal definition of \query}
\label{sec:formal}

Randomness has long been used in mathematical algorithms, and its 
formal role in computations dates to shortly after the introduction of
Turing machines.
In his paper \cite{Tur50} introducing the Turing test,
Turing informally discussed introducing a ``random element'',
and in a radio discussion c.\ 1951 (later published as \cite{Tur96}),
he considered placing a random string of $0$'s and $1$'s on an additional
input bit tape of a Turing machine.
In 1956, de~Leeuw, Moore, Shannon, and Shapiro \cite{dMSS56} proposed
probabilistic Turing machines (PTMs) more formally, making use 
of Turing's formalism \cite{Tur39} for oracle Turing machines:
a PTM is an oracle Turing machine whose oracle tape
comprises independent random bits.  From this perspective, the output of a PTM is itself a random variable and so we may speak of \emph{the distribution of (the output of) a PTM}.  For the PTM \query, which simulates other PTMs passed as inputs, we can express its distribution in terms of the distributions of PTM inputs.  In the remainder of this section, we 
describe this formal framework and then use it to explore the class of distributions that may be represented by PTMs.

Fix a
canonical enumeration of (oracle) Turing machines and the corresponding
partial computable (oracle) functions $\{\varphi_e\}_{e\in\Naturals}$,
each considered as a partial function 
\begin{align*}
    \{0,1\}^\infty \times \{0,1\}^* \to \{0,1\}^*,
\end{align*}
where  $\{0,1\}^\infty$ denotes the set of countably infinite binary strings and, as before, $\{0,1\}^*$
denotes the set of finite binary strings.
One may think of each such partial function as a mapping from an oracle tape and input tape to an output tape.  We will write $\varphi_e(x,s)\!\downarrow$ when $\varphi_e$ is defined on oracle tape $x$ and input string $s$, and $\varphi_e(x,s)\!\uparrow$ otherwise.  We will write $\varphi_e(x)$ when the input string is empty or when there is no input tape.
As a model for the random bit tape,
we define an independent and identically distributed (\iid)\ sequence $R=(R_i : i \in \Nats)$ of binary random variables, each taking the value $0$ and $1$ with equal probability, i.e, each $R_i$ is an independent $\BernoulliD(1/ 2)$ random variable. 
We will write $\Pr$ to denote the distribution of the random bit tape $R$.
More formally, $R$ will be considered to be the identity function on the Borel probability space
$(\{0,1\}^\infty, \Pr)$, where $\Pr$ is the countable product of $\BernoulliD(1/2)$ measures.

Let $s$ be a finite string, let $e \in\Nats$, and suppose
that 
\begin{align*}
\Pr \{\, r \in \{0,1\}^\infty \st \varphi_e(r,s)\!\downarrow \,\} = 1.
\end{align*}
Informally, we will say that the probabilistic Turing machine (indexed by) $e$ halts almost surely on input $s$.
In this case, we define the \defn{output distribution of the $e$th (oracle) Turing
machine on input string $s$} to be the distribution of the random variable
\begin{align*}
     \varphi_e(R,s);
\end{align*}
we may directly express this distribution as
\begin{align*}
    \Pr \circ \varphi_e( \pars, s)^{-1}.
\end{align*}

Using these ideas we can now formalize \query.
In this formalization, both the prior and predicate programs $\PP$ and $\CC$ passed as input to \query\ are finite binary strings interpreted as indices for a probabilistic Turing machine with no input tape.  
Suppose that \PP\ and \CC\ halt almost surely.
In this case, the output distribution of $\query(\PP,\CC)$ can be characterized as follows:
Let $R=(R_i \st i \in \Nats)$ denote the random bit tape, 
let $\pi : \Nats \times \Nats \to \Nats$ be a standard pairing function (i.e., a computable bijection), and,
for each $n,i \in \Nats$, let $R^{(n)}_i \defas R_{\pi(n,i)}$ so that $\{R^{(n)} \st n \in \Nats \}$ are
independent random bit tapes, each with distribution $\Pr$.  
Define the $n$th \defn{sample from the prior} to be the random variable
\begin{align*}
     X_n \defas \varphi_{\PP}(R^{(n)}),
\end{align*}
and let 
\begin{align*}
N \defas \inf\, \{\, n \in \Nats \st \varphi_{\CC}(R^{(n)}) = 1\, \}
\end{align*}
be the first iteration $n$ such that the predicate $\CC$ evaluates to 1 (i.e., accepts).  The \defn{output distribution of $\query(\PP,\CC)$} is then the distribution of the random variable
\begin{align*}
X_{N},
\end{align*}
whenever $N < \infty$ holds with probability one, and is undefined otherwise.  
Note that $N < \infty$ \as\ if and only if $\CC$ accepts with non-zero probability.
As above, we can give a more direct characterization: Let 
\begin{align*}
\mathcal A \defas \{\, R \in \{0,1\}^\infty \st   \varphi_{\CC}(R) = 1\, \}
\end{align*}
be the set of random bit tapes $R$ such that the predicate $\CC$ accepts by outputting $1$.  The condition
``$N < \infty$ with probability one'' is then equivalent to the statement that $\Pr(\mathcal A) > 0$. In that case, we may express the output distribution of $\query(\PP,\CC)$ as 
\begin{align*}
\Pr_{\mathcal A} \circ \varphi_{\PP}^{-1}
\end{align*}
where $\Pr_{\mathcal A}(\cdot ) \defas \Pr(\,\cdot \given {\mathcal A})$ is the distribution of the random bit tape conditioned on $\CC$ accepting (i.e., conditioned on the event ${\mathcal A}$).

\subsection{Computable measures and probability theory}

Which probability distributions are the output distributions of \emph{some} PTM? 
In order to investigate this question, consider what we might learn from
simulating a given PTM $\PP$ (on a particular input) that halts almost surely.  More precisely, for a finite bit string $r \in \{0,1\}^*$ with length $|r|$, consider simulating $\PP$, replacing its random bit tape with the finite string $r$: If, in the course of the simulation, the program attempts to read beyond the end of the finite string $r$, we terminate the simulation prematurely.  On the other hand, if the program halts and outputs a string $t$ then we may conclude that all simulations of $\PP$ will return the same value when the random bit tape begins with $r$.  As the set of random bit tapes beginning with $r$ has $\Pr$-probability $2^{-|r|}$, we may conclude that the distribution of $\PP$ assigns at least this much probability to the string $t$.

It should be clear that, using the above idea, we may enumerate the (prefix-free) set of strings $\{r_n\}$, and matching outputs $\{t_n\}$, such that $\PP$ outputs $t_n$ when its random bit tape begins with $r_n$.  It follows that, for all strings $t$ and $m \in \Nats$,
\begin{align*}
\sum_{\{n \le m \st t_n = t\}} 2^{-|r_n|} 
\end{align*}
is a lower bound on the probability that the distribution of $\PP$ assigns to $t$, and
\begin{align*}
1-\sum_{\{n \le m \st t_n \neq t\}} 2^{-|r_n|}
\end{align*}
is an upper bound.  Moreover, it is straightforward to show that as $m\to\infty$, these converge monotonically from above and below to the probability that $\PP$ assigns to the string $t$.

This sort of effective information about a real number precisely
characterizes the \emph{computable real numbers}, first described by
Turing in his paper \cite{Tur36} introducing Turing machines.
For more details, see the survey by Avigad and Brattka connecting computable analysis to work of Turing, elsewhere in this volume \cite{AB13}.

\begin{definition}[computable real number]
A real number $r\in\Reals$ is said to be \defn{computable} when its left and right cuts of
rationals 
$\{q\in\Rationals \st q<r\},  
\{q\in\Rationals \st r<q\} $  
are computable (under the canonical computable encoding of rationals).
Equivalently, a real is computable when there is a computable sequence of
rationals $\{q_n\}_{n\in\Naturals}$ that \emph{rapidly converges to $r$}, in the sense
that
$|q_n - r| < 2^{-n}$ for each $n$.
\end{definition}

We now know that the probability of each output string $t$ from a PTM is a computable real (in fact, \emph{uniformly in $t$}, i.e., this probability can be computed for each $t$ by a single program that accepts $t$ as input.). Conversely, for every computable real $\alpha \in [0,1]$ and string $t$, there is a PTM that outputs $t$ with probability $\alpha$.  In particular, let $R=(R_1,R_2,\dotsc)$ be our random bit tape, let $\alpha_1,\alpha_2,\dotsc$ be a uniformly computable sequence of rationals that rapidly converges to $\alpha$, and consider the following simple program: On step $n$, compute the rational $A_n\defas \sum_{i=1}^n R_i \cdot 2^{-i}$.
\linebreak If $A_n < \alpha_n - 2^{-n}$, then halt and output $t$; If
$A_n > \alpha_n + 2^{-n}$, then halt and output $t0$.  Otherwise, proceed to step $n+1$. 
Note that $A_\infty \defas \lim A_n$ is uniformly distributed in the unit interval, and so $A_\infty < \alpha$ with probability $\alpha$.  Because $\lim \alpha_n \to \alpha$, the program eventually halts for all but one (or two, in the case that $\alpha$ is a dyadic rational) random bit tapes.  In particular, if the random bit tape is the binary expansion of $\alpha$, or equivalently, if $A_\infty = \alpha$, then the program does not halt, but this is a $\Pr$-measure zero event.

Recall that we assumed, in defining the output distribution of a PTM, that the program
halted almost surely.  The above construction illustrates why the stricter requirement that PTMs halt always (and not just almost surely) could be very limiting. In fact, one can show that there is no PTM that halts always and whose output distribution assigns, e.g., probability $1/3$ to 1 and $2/3$ to 0.
Indeed, the same is true for all non-dyadic probability values (for details see \cite[Prop.~9]{AFR11}).

We can use this construction to sample from any distribution $\nu$ on $\{0,1\}^*$ for which we can compute the probability of a string $t$ in a uniform way.  In particular, fix an enumeration of all strings $\{t_n\}$ and, for each $n \in \Nats$, define the distribution $\nu_n$ on $\{t_n,t_{n+1},\dotsc\}$ by $\nu_n = \nu / (1 - \nu \{ t_1,\dotsc, t_{n-1}\})$.  If $\nu$ is computable in the sense that for any $t$, we may compute real $\nu\{t\}$ uniformly in $t$, then $\nu_n$ is clearly computable in the same sense, uniformly in $n$.  We may then proceed in order, deciding whether to output $t_n$ (with probability $\nu_n\{t_n\}$) or to recurse and consider $t_{n+1}$.  It is straightforward to verify that the above procedure outputs a string $t$ with probability $\nu\{t\}$, as desired.

These observations motivate 
the following definition of a computable probability measure, 
which is a special case of notions from computable analysis developed
later; for details of the history see \cite[\S1]{Wei99}.

\begin{definition}[computable probability measure]
A probability measure on $\{0,1\}^*$ is said to be \defn{computable}  when 
the measure of each string is a computable real, uniformly in the string.
\end{definition}

The above argument demonstrates that the 
samplable probability measures --- those distributions on
$\{0,1\}^*$ that arise from sampling procedures performed by probabilistic Turing
machines that halt a.s.\ --- coincide with computable probability measures.

While in this paper we will not consider the efficiency of these
procedures, it is worth noting that while the class of distributions that
can be
sampled by Turing machines coincides with the class of computable probability measures on $\{0,1\}^*$, the analogous
statements for polynomial-time Turing machines fail. In particular, there are distributions from
which one can efficiently sample, but for which output probabilities are
not efficiently computable (unless $\textbf{P} = \textbf{PP}$), for
suitable formalizations of these concepts \cite{Yam99}.

\subsection{Computable probability measures on uncountable spaces}
\label{compprobmeas}
So far we have considered distributions on the space of finite
binary strings. 
Under a suitable encoding, PTMs can be seen to represent distributions on general countable
spaces.   On the other hand, many phenomena are naturally modeled in terms of continuous quantities like real numbers. 
In this section we will look at the problem of representing distributions 
on uncountable spaces, and then consider the problem of extending \query\ in a similar direction.

To begin, we will describe distributions on the space of
\emph{infinite} binary strings, $\{0,1\}^\infty$.  Perhaps the most natural proposal for representing such distributions is to again consider PTMs whose output can be interpreted as representing a random point in $\{0,1\}^\infty$.  As we will see, such distributions will have an equivalent characterization in terms of uniform computability of the measure of a certain class of sets.

Fix a computable bijection between $\Nats$ and finite binary strings,
and for $n \in \Nats$, write $\bar n$ for the image of $n$ under this map.
Let $e$ be the index of some PTM, and suppose that $\varphi_e(R,\bar n)\in\{0,1\}^n$ and $\varphi_e(R,\bar n)
\sqsubseteq \varphi_e(R,\overline {n+1})$ almost surely for all $n\in\Nats$, where $r \sqsubseteq s$ for two
binary strings $r$ and $s$ when $r$ is a prefix of $s$.  Then the \emph{random point in $\{0,1\}^\infty$ given by $e$} is defined to be 
\begin{align}\label{limobj}
\lim_{n\to\infty} (\varphi_e(R,\bar n),0,0,\dotsc).
\end{align}
Intuitively, we have represented the (random) infinite object by a program (relative to a fixed random bit tape) that can provide a convergent sequence of finite approximations.

It is obvious that the distribution of $\varphi_e(R,\bar n)$ is computable, uniformly in $n$.  As a consequence, for every basic clopen set $A=\{ s \st r \sqsubseteq s \}$, we may compute the probability that the limiting object defined by \eqref{limobj} falls into $A$, and thus we may compute arbitrarily good lower bounds for the measure of unions of computable sequences of basic clopen sets, i.e., c.e.\ open sets.

This notion of computability of a measure is precisely that developed in
computable analysis, and in particular, via
the Type-Two Theory of Effectivity
(TTE); for details see Edalat \cite{Eda96}, Weihrauch \cite{Wei99}, 
Schr\"oder \cite{Sch07}, 
and G\'acs \cite{Gac05}.
This formalism rests on Turing's oracle machines \cite{Tur39}; for more details, again see the survey by Avigad and Brattka elsewhere in this volume \cite{AB13}.
The representation of a measure by the values assigned to basic clopen
sets can be interpreted in several ways, each of which allows us to
place measures on spaces other than just the set of infinite strings.
From a topological point of view,
the above representation involves the choice of a particular basis for the
topology, with an appropriate enumeration, making 
$\{0,1\}^\infty$ into a \emph{computable topological space}; for details, see 
\cite[Def.~3.2.1]{Wei00} and \cite[Def.~3.1]{GSW07}.

Another approach is to place a metric on $\{0,1\}^\infty$ that induces the
same topology, and that is computable on a dense set of points, making it
into a \emph{computable metric space}; see 
 \cite{Hem02} and \cite{Wei93}  on approaches in TTE,
\cite{Bla97} and \cite{EH98} in effective domain theory, and
\cite[Ch.~8.1]{Wei00} and \cite[{\S}B.3]{Gac05} for more details.
For example, one could have defined the distance between two strings in
$\{0,1\}^\infty$ to be $2^{-n}$, where $n$ is the location of the first bit
on which they differ; instead choosing $1/n$ would have given a different
metric space but would induce the same topology, and hence the same notion
of computable measure.
Here we use the following definition of a computable metric space, taken
from \cite[Def.~2.3.1]{GHR10}.
\begin{definition}[computable metric space]
A \defn{computable metric space} is a triple \linebreak $(S,\delta,\cD)$ for which
$\delta$ is a metric on the set $S$ satisfying
\begin{enumerate}
\item $(S,\delta)$ is a complete separable metric space;
\item $\cD = \{s(1),s(2),\dotsc\}$ is an enumeration of a dense subset
of $S$; and,
\item the real numbers $\delta(s(i),s(j))$ are computable, uniformly in $i$
and $j$.
\end{enumerate}
\end{definition}

We say that an $S$-valued random variable 
$X$ (defined on the same space as $R$) is an \defn{(almost-everywhere) computable $S$-valued random variable} or \defn{random point in $S$}  when there is a PTM $e$ such that $\delta(X_n,X) < 2^{-n}$ almost surely for all $n \in \Nats$, where $X_n \defas s(\varphi_e(R,\bar n))$.
We can think of the random sequence $\{X_n\}$ as a \defn{representation} of the
random point $X$.
A \defn{computable probability measure} on $S$ is precisely the distribution of
such a random variable.

For example, the real numbers form a computable metric space $(\Reals, d,
\Rationals)$, where $d$ is the Euclidean metric, and $\Rationals$ has the
standard enumeration.  One can show that computable probability measures on
$\Reals$ are then those for which the measure of an arbitrary finite union
of rational open intervals admits arbitrarily good lower bounds,
uniformly in (an encoding of) the sequence of intervals.  Alternatively, one can show that the space of
probability measures on $\Reals$ is a computable metric space under the Prokhorov metric, with respect to (a
standard enumeration of) a dense set of atomic measures with finite support in the rationals.  The notions
of computability one gets in these settings align with classical notions. For example, the set of naturals
and the set of finite binary strings are indeed both computable metric spaces, and the computable measures
in this perspective are precisely as described above.

Similarly to the countable case, we can use \query\ to sample points in \emph{uncountable spaces} conditioned on a predicate.  Namely, suppose the prior program \PP\ represents a random point in an
uncountable space with distribution $\nu$.  For any string $s$, write $\PP(s)$ for \PP\ with the input fixed to $s$, and let \CC\ be
a predicate that accepts with non-zero probability.  
Then the PTM that, on input  $\bar n$, outputs the result of simulating $\query(\PP(\bar n),\CC)$ 
is a representation of $\nu$ conditioned on
the predicate accepting. When convenient and clear from context, we will denote this derived PTM by simply writing $\query(\PP, \CC)$.

\subsection{Conditioning on the value of continuous random variables}

The above use of \query\ allows us to condition a model of a computable real-valued random variable $X$ on a predicate $\CC$. 
However, the restriction on predicates (to accept with non-zero probability) and the definition of \query\ itself do not, in general, allow us to condition on $X$ itself taking a specific value.  Unfortunately, the problem is not superficial, as we will now relate.  

Assume, for simplicity, that $X$ is also continuous (i.e., $\Pr\{X=x\} = 0$ for all reals $x$).  
Let $x$ be a computable real, and for every computable real $\varepsilon>0$, consider the (partial computable) predicate $\CC_{\varepsilon}$ that accepts when $|X-x| < \varepsilon$, rejects when \linebreak $|X-x|> \varepsilon$, and is undefined otherwise.  
(We say that such a predicate \emph{almost} decides the event $\{|X-x|<\varepsilon\}$ as it decides the set outside a measure zero set.)  
We can think of
$
\query(\PP,\CC_{\varepsilon})
$
as a ``positive-measure approximation'' to conditioning on $X=x$.
Indeed, if \PP\ is a prior program that samples a computable random variable $Y$ and
$B_{x,\varepsilon}$ denotes the closed $\varepsilon$-ball around $x$, then this \query\ corresponds to the
conditioned
distribution $\Pr(Y\given X\in B_{x,\varepsilon})$, and so provided $\Pr\{X \in B_{x,\varepsilon}\} > 0$, this is well-defined and evidently computable.
But what is its relationship to the original problem?

While one might be inclined to think that 
$
\query(\PP,\CC_{\varepsilon=0})
$
represents our original goal of conditioning on $X=x$, the continuity of the random variable $X$ implies that $\Pr\{X \in B_{x,0}\} = \Pr\{X=x\} = 0$ and so $\CC_0$ rejects with probability one.  It follows that $\query(\PP,\CC_{\varepsilon=0})$ does not halt on any input, and thus does not represent a distribution.   

The underlying problem is that, in general, conditioning on a null set is mathematically undefined.
The standard measure-theoretic solution is to consider the so-called ``regular conditional distribution'' given by conditioning
on the $\sigma$-algebra generated by $X$---but even this approach would in general fail to solve our problem because the resulting disintegration is only defined up to a null set, and so is undefined at points (including $x$).
(For more details, see \cite[\S{III}]{AFR11} and 
\cite[Ch.~9]{Tju80}.)

There have been various attempts at more
constructive approaches, e.g.,
Tjur \cite{Tju74, Tju75, Tju80}, Pfanzagl
\cite{Pfa79}, and
Rao \cite{Rao88, Rao05}.  One approach worth highlighting is due to Tjur \cite{Tju75}.  There
he considers additional hypotheses that are equivalent to the existence of a continuous \emph{disintegration}, which must then be unique at all points.  (We will implicitly use this notion henceforth.)
Given the connection between computability and continuity, a natural question to ask is whether we might be able to extend \query\ along the lines.

Despite various constructive efforts, no general method had been found for computing conditional distributions.
In fact, conditional distributions are not in general computable, as shown
by Ackerman, Freer, and Roy \cite[Thm.~29]{AFR11},
and
it is for this reason we have defined \query\ in terms of conditioning on the event $\CC=1$, which, provided that $\CC$ accepts with non-zero probability as we have required, is a positive-measure event.
The proof of the noncomputability of conditional probability
\cite[\S{VI}]{AFR11}
involves an encoding of the halting problem into a pair $(X,Y)$ of computable (even, absolutely continuous) random variables
 in $[0,1]$ such that no ``version'' of the conditional distribution $\Pr(Y
\given X = x)$ is a computable function \linebreak of $x$.

What, then, is the relationship between conditioning on $X=x$ and the approximations $\CC_\varepsilon$ defined above?
In sufficiently nice settings, the distribution represented by $\query(\PP, \CC_{\varepsilon})$ converges to the desired distribution as $\varepsilon \to 0$.
But as a corollary of the aforementioned noncomputability result,
one sees that it is noncomputable in general to determine a value of $\varepsilon$ from a desired level of accuracy to the desired distribution, for if there were such a general and computable relationship, one could use it to compute conditional distributions, a contradiction.
Hence although such a sequence of approximations might converge in
the limit, one cannot in general compute how close it is to convergence.

On the other hand, the presence of noise in measurements can lead to computability.
As an example, consider the problem of representing the distribution of $Y$ conditioned on $X + \xi = x$, 
where $Y$, $X$, and $x$ are as above, and $\xi$ is independent of $X$ and $Y$ and uniformly distributed on the interval $[-\varepsilon, \varepsilon]$.  
While conditioning on continuous random variables is not computable in general, here it is possible.  
In particular, note that  $\Pr(Y \given X + \xi = x) = \Pr(Y \given X \in B_{x,\varepsilon})$ and so $\query(\PP,\CC_\varepsilon)$ represents the desired distribution.

This example can be generalized considerably beyond uniform noise (see
\cite[Cor.~36]{AFR11}).   Many models considered in practice posit the existence
of independent noise in the quantities being measured, and so the \query\ formalism can be used to capture
probabilistic reasoning in these settings as well.  However, in general we should not expect to be able to reliably approximate noiseless measurements with noisy measurements, lest we contradict the noncomputability of conditioning.  Finally, it is important to note that the computability that arises in the case of certain types of independent noise is a special case of the computability that arises from the existence and computability of certain conditional probability densities \cite[\S{VII}]{AFR11}. This final case covers most models that arise in statistical practice, especially those that are finite-dimensional.

In conclusion, while we cannot hope to condition on arbitrary computable random variables,
\query\ covers nearly all of the situations that arise in practice, and suffices for our
purposes.
Having laid the theoretical foundation for \query\ and described its connection with conditioning, 
we now return to the medical diagnosis example and more elaborate uses of \query,
with a goal of understanding additional features of the formalism.

\section{Conditional independence and compact representations}
\label{sec:condindep}

In this section, we return to the medical diagnosis example,
and explain the way in which conditional independence leads to compact representations, and conversely, the fact that efficient probabilistic programs, like \DS, exhibit many conditional independencies. 
We will do so through connections with the 
Bayesian network formalism, whose introduction by Pearl \cite{Pea88} was a major advancement in AI. 

\subsection{The combinatorics of \query}

Humans engaging in common-sense reasoning often seem to possess an
unbounded range of responses or behaviors;
this is perhaps unsurprising given the enormous 
variety of possible situations that can arise, even in simple domains.

Indeed, the small handful of potential diseases and symptoms that our medical diagnosis model posits already gives rise to a combinatorial explosion of potential scenarios with which a doctor could be faced: 
among 11 potential diseases and 7 potential symptoms there are 
\begin{align*}
3^{11}\cdot3^{7} = 387\,420\,489
\end{align*}
partial assignments to a subset of variables. 

Building a table (i.e., function) associating every possible diagnostic scenario with a response would be an extremely difficult task, and probably nearly impossible if one did not take advantage of  some structure in the domain to devise a more compact representation of the table than a structureless, huge list.   
In fact, much of AI can be interpreted as proposals for specific structural assumptions that lead to more compact representations, and the \query\ framework can be viewed from this perspective as well: 
the prior program \DS\  implicitly defines a full table of responses, and the predicate can be understood as a way to index into this vast table.

This leads us to three questions:  
Is the table of diagnostic responses induced by \DS\ any good?
How is it possible that so many responses can be encoded so compactly? And what properties of a model follow from the existence of an efficient prior program, as in the case of our medical diagnosis example and the prior program \DS?
In the remainder of the section we will address the latter two questions, returning to the former in Section~\ref{sec:latent} and Section~\ref{sec:definetti}.

\subsection{Conditional independence}
\label{subsec:condindependence}

Like \DS, every probability model of $18$ binary variables implicitly defines a gargantuan set of conditional probabilities.  
However, unlike \DS, most such models have no compact representation.  
To see this, note that a probability distribution over $k$ outcomes is, in general, specified by $k-1$ probabilities, and so in principle, in order to specify a distribution on $\{0,1\}^{18}$, one must specify
\begin{align*}
2^{18} - 1 = 262\,143
\end{align*}
probabilities.  
Even if we discretize the probabilities to some fixed accuracy, a simple counting argument shows that most such distributions have no short description.

In contrast, 
Table~\ref{tab:probs} contains only
\begin{align*}
11 + 7 + 11\cdot 7 = 95 
\end{align*}
probabilities, which, via the small collection of probabilistic computations performed by \DS\ and described informally in the text, parameterize a distribution over $2^{18}$ possible outcomes.  
What properties of a model can lead to a compact representation?

The answer to this question is \emph{conditional independence}.  Recall that a collection of random
variables $\{X_i \st i \in I\}$ is \defn{independent} when, for all finite subsets $J \subseteq I$ and measurable
sets $A_i$ where $i \in J$, we have
\begin{align}\label{indfact}
\Pr \bigl( \bigwedge_{i\in J} X_i\in A_i \bigr) = \prod_{i \in J} \Pr(X_i \in A_i).
\end{align}
If $X$ and $Y$ were binary random variables, then specifying their distribution would require $3$ probabilities in general, but only $2$ if they were independent.  
While those savings are small, consider instead $n$ binary random variables $X_j$, $j=1,\dotsc, n$, and note that, 
while a generic distribution over these random variables would require the specification of $2^n-1$ probabilities, only $n$ probabilities are needed in the case of full independence.

Most interesting probabilistic models with compact representations will not exhibit enough independence between their constituent random variables to explain their own compactness in terms of the factorization in \eqref{indfact}.
Instead, the slightly weaker (but arguably more fundamental) notion of conditional independence is needed.  
Rather than present the definition of conditional independence in its full generality, we will consider a special case, restricting our attention to conditional independence with respect to a discrete random variable $N$ taking values in some countable or finite set $\mathcal N$.  
(For the general case, see Kallenberg \cite[Ch.~6]{Kal02}.)
We say that a collection of random variables $\{X_i \st i \in I\}$ is \defn{conditionally independent} 
given $N$ when, for all $n \in \mathcal N$, finite subsets $J \subseteq I$ and
measurable sets $A_i$, for $i \in J$, we have
\begin{align*}
\Pr \bigl( \bigwedge_{i\in J} X_i\in A_i 
\given N = n\bigl) = \prod_{i \in J} \Pr(X_i \in A_i \given N = n).
\end{align*}
To illustrate the potential savings that can arise from conditional independence, consider $n$ binary random variables that are conditionally independent given a discrete random variable taking $k$ values.  
In general, the joint distribution over these $n+1$ variables is specified by $k\cdot2^n - 1 $ probabilities, but, in light of the conditional independence, we need specify only 
$k (n+1) - 1$ probabilities.

\subsection{Conditional independencies in \DS}
\label{subsec:condindepinDS}

In Section~\ref{subsec:condindependence},
we saw that conditional independence gives rise to compact representations.
As we will see, the variables in \DS\ exhibit many conditional independencies.

To begin to understand the compactness of \DS, note that the 95 variables
\begin{align*}
\{ D_1,\dotsc,D_{11};\, L_1,\dotsc,L_{7};\, C_{1,1},C_{1,2},C_{2,1},C_{2,2},\dotsc,C_{11,7}\}
\end{align*}
are independent, and thus their joint distribution is determined by specifying only $95$ probabilities (in particular, those in Table~\ref{tab:probs}).  Each symptom $S_m$ is then derived as a deterministic function of a 23-variable subset 
\begin{align*}\{D_1, \dotsc, D_{11};\, L_m;\, C_{1,m}, \dotsc, C_{11,m} \},\end{align*} which implies that the symptoms are conditionally independent given the diseases.
However, these facts alone do not fully explain the compactness of \DS.\ \  In particular, there are
\begin{align*}
2^{2^{23}} > 10^{10^6}
\end{align*}
binary functions of $23$ binary inputs, and so by a counting argument, most have no short description.  
On the other hand, the $\max$ operation that defines $S_m$ does have a compact \emph{and efficient} implementation.
In Section~\ref{sec:effrep} we will see that this implies that we can introduce
additional random variables representing intermediate quantities produced in the process of computing each symptom $S_m$ from its corresponding collection of 23-variable ``parent'' variables, and that these random variables exhibit many more conditional independencies than exist between $S_m$ and its parents.
From this perspective, the compactness of \DS\ is tantamount to there being only a small number of such variables that need to be introduced.
In order to simplify our explanation of this connection, we pause to introduce the idea of representing conditional independencies using graphs.

\subsection{Representations of conditional independence}
\label{sec:bayesnets}

\begin{figure}
\centering
 \begin{center}

\def\cdone#1#2{
  \node [right] at (2.5-2.5*#2*#2*#2/144/12,11-#2) {\ $C_{#2,#1}$};
  \node           at (2.5-2.5*#2*#2*#2/144/12,11-#2) {$\node$};
  \draw (0,11) -- (2.5-2.5*#2*#2*#2/144/12,11-#2); 
}

\def\symptom#1#2%
{
  \begin{scope}[#2]
  \node [left] at (0,11) {$S_{#1}$\ \ };
  \node         at (0,11) {$\node$};
  \node [right] at (0,-1) {\ $L_{#1}$\ \ };
  \node           at (0,-1) {$\node$};
  \draw (0,11) -- (0,-1); 

  \cdone{#1}{11}
  \cdone{#1}{10}
  \cdone{#1}{9}
  \cdone{#1}{8}
  \cdone{#1}{7}
  \cdone{#1}{6}
  \cdone{#1}{5}
  \cdone{#1}{4}
  \cdone{#1}{3}
  \cdone{#1}{2}
  \cdone{#1}{1}
  \end{scope}
}

\def\dslink#1#2{
  \draw (5*#2-5,-13) -- (3*#1+1,0);
}

\def\disease#1#2%
{
  \node [above] at (3*#1+1,.5) {$D_{#1}$};
  \node at (3*#1+1,0) {$\node$};
  \dslink{#1}{1}
  \dslink{#1}{2}
  \dslink{#1}{3}
  \dslink{#1}{4}
  \dslink{#1}{5}
  \dslink{#1}{6}
  \dslink{#1}{7}
}

\scalebox{.5}{
\begin{tikzpicture}[>=stealth,scale=.5]
  \symptom{1}{xshift=0,yshift=0}
  \symptom{2}{xshift=5cm,yshift=0}
  \symptom{3}{xshift=10cm,yshift=0}
  \symptom{4}{xshift=15cm,yshift=0}
  \symptom{5}{xshift=20cm,yshift=0}
  \symptom{6}{xshift=25cm,yshift=0}
  \symptom{7}{xshift=30cm,yshift=0}
  \begin{scope}[xshift=0cm,yshift=24cm]
   \disease{1}{2}
   \disease{2}{2}
   \disease{3}{2}
   \disease{4}{2}
   \disease{5}{2}
   \disease{6}{2}
   \disease{7}{2}
   \disease{8}{2}
   \disease{9}{2}
   \disease{10}{2}
   \disease{11}{2}
  \end{scope}

\end{tikzpicture}
}%
 \end{center}
\caption{Directed graphical model representations of the conditional independence underlying the medical diagnosis example.  (Note that the directionality of the arrows has not been rendered as they all simply point towards the symptoms $S_m$.) }
\label{fig:gms}
\vspace{5mm}
\begin{center}

\scalebox{1}{
\begin{tikzpicture}[>=stealth,scale=.5,decoration={markings,mark=at position 0.75 with {\arrow{>}}}]
\begin{scope}
  \node at (2,-2) {$\node$};
  \node at (2,-1) {$L_m$};
  \node at (2,0) {$\node$};
  \node at (2,1) {$C_{n,m}$};
  \node at (0,-2) {$\node$};
  \node at (0,-3) {$S_m$};
  \node at (0,2) {$\node$};
  \node at (0,3) {$D_n$};
  \draw (-1,1.5) -- (4,1.5) -- (4,-4) -- (-1,-4) -- (-1,1.5);
  \node[font=\tiny] [above left] at (4,-4) {symptoms $m$\ };
  \draw (-1.5,3.5) -- (3.5,3.5) -- (3.5,-0.5) -- (-1.5,-0.5) -- (-1.5,3.5);
  \node[font=\tiny] [below left] at (3.5,3.5) {diseases $n$\ };
  \draw[postaction={decorate}] (0,2) -- (0,-2); 
  \draw[postaction={decorate}] (2,0) -- (0,-2); 
  \draw[postaction={decorate}] (2,-2) -- (0,-2); 
\end{scope}
\end{tikzpicture}
}
  \end{center}
\caption{The repetitive structure Figure~\ref{fig:gms} can be partially captured by so-called ``plate notation'', which can be interpreted as a primitive {\tt for}-loop construct.  Practitioners have adopted a number of strategies like plate notation for capturing complicated structures.}
\label{fig:gms2}
\end{figure}

A useful way to represent conditional independence among a collection of random variables is in terms of a directed acyclic graph, where the vertices stand for random variables, and the collection of edges indicates the presence of certain conditional independencies.  
An example of such a graph, known as a directed graphical model or Bayesian network, is given in Figure~\ref{fig:gms}.  
(For more details on Bayesian networks, see the survey by Pearl \cite{Pea04}.  
It is interesting to note that Pearl cites Good's ``causal calculus''
\cite{Goo61}---which we have already encountered in connection with our
medical diagnosis example, and which was based in part on Good's wartime
work with Turing on the weight of evidence---as a historical antecedent to
Bayesian networks \cite[\S70.2]{Pea04}.)

Directed graphical models often capture the ``generative'' structure of a collection of random variables: informally, by the direction of arrows, the diagram captures, for each random variable, which other random variables were directly implicated in the computation that led to it being sampled.  
In order to understand exactly which conditional independencies are formally encoded in such a graph, we must introduce the notion of $d$-separation.

We determine whether a pair $(x,y)$ of vertices are $d$-separated by a subset of vertices $\mathcal E$ as follows:  
First, mark each vertex in $\mathcal E$ with a $\times$, which we will indicate by the symbol $\enode$.  
If a vertex with (any type of) mark has an unmarked parent, mark the parent with a $+$, which we will indicate by the symbol $\anode$.   
Repeat until a fixed point is reached.  
Let $\node$ indicate unmarked vertices.  
Then $x$ and $y$ are $d$-separated if, for all (undirected) paths from $x$ to $y$ through the graph, one of the following patterns appears:
\begin{align*}
\textstyle
\rightarrow \enode \rightarrow \\
\textstyle
\leftarrow \enode \leftarrow \\
\textstyle
\leftarrow \enode \rightarrow \\
\textstyle
\rightarrow \node \leftarrow 
\end{align*}
More generally, if $\mathcal X$ and $\mathcal E$ are disjoint sets of vertices, then the graph encodes the conditional independence of the vertices $\mathcal X$ given $\mathcal E$ if every pair of vertices in $\mathcal X$ is $d$-separated given $\mathcal E$.  
If we fix a collection $V$ of random variables, then we say that a directed acyclic graph $G$ over $V$ is a \emph{Bayesian network} (equivalently, a directed graphical model) when the random variables in $V$ indeed posses all of the conditional independencies implied by the graph by $d$-separation.  
Note that a directed graph $G$ says nothing about which conditional independencies do \emph{not} exist among its vertex set.

Using the notion of $d$-separation, we can determine from the Bayesian network in Figure~\ref{fig:gms} that the diseases $\{D_1,\dotsc,D_{11}\}$ are independent (i.e., conditionally independent given $\mathcal E = \emptyset$).  
We may also conclude that the symptoms $\{S_1,\dotsc,S_7\}$ are conditionally independent given the diseases $\{D_1,\dotsc,D_{11}\}$. 

In addition to encoding a set of conditional independence statements that hold among its vertex set, directed graphical models demonstrate that the joint distribution over its vertex set admits a concise factorization:  
For a collection of binary random variables $X_1,\dotsc,X_k$, write $p(X_1,\dotsc,X_k) : \{0,1\}^k \to [0,1]$ for the probability mass function (p.m.f.) taking an assignment $x_1,\dotsc,x_k$ to its probability $\Pr( X_1 = x_1, \dotsc, X_k=x_k)$, and write 
\begin{align*}p(X_1,\dotsc,X_k \given Y_1,\dotsc,Y_m) : \{0,1\}^{k+m} \to [0,1]\end{align*} for the \emph{conditional} p.m.f.\
corresponding to the conditional distribution 
\begin{align*}\Pr(X_1,\dotsc,X_k \given Y_1,\dotsc,Y_m).\end{align*}  It is a basic fact from probability that 
\begin{align}\label{useless}
p(X_1,\dotsc,X_k) 
&= p(X_1) \cdot p(X_2\given X_1) \dotsm p(X_k \given X_1,\dotsc,X_{k-1})\\
&= \prod_{i=1}^k p(X_i \given X_j,\ j < i).
\nonumber
\end{align}
Such a factorization provides no advantage when seeking a compact representation, 
as a conditional p.m.f.\ of the form $p(X_1,\dotsc,X_k \given Y_1,\dotsc,X_m)$ is determined by $2^m \cdot (2^k-1)$ probabilities.  
On the other hand, if we have a directed graphical model over the same variables, then we may have a much more concise factorization.  
In particular, let $G$ be a directed graphical model over $\{X_1,\dotsc,X_k\}$, and write $\PA(X_j)$ for the set of vertices $X_i$ such that $(X_i,X_j) \in G$, i.e., $\PA(X_j)$ are the parent vertices of $X_j$.  
Then the joint p.m.f.\ may be expressed as
\begin{align}\label{gmfactor}
p(X_1,\dotsc,X_k) = \prod_{i=1}^k p(X_i \given \PA(X_i) ).
\end{align}
Whereas the factorization given by \eqref{useless} requires the full set of $\sum_{i=1}^k 2^{i-1} = 2^k-1$ probabilities to determine, this factorization requires $\sum_{i=1}^k 2^{|\PA(X_i)|}$ probabilities, which in general can be exponentially smaller in $k$.

\subsection{Efficient representations and conditional independence}
\label{sec:effrep}

As we saw at the beginning of this section,
models with only a moderate number of variables can have enormous descriptions.
Having introduced the directed graphical model formalism, 
we can use \DS\ as an example to explain why, roughly speaking, the output distributions of efficient probabilistic programs exhibit many conditional independencies.

What does the efficiency of \DS\ imply about the structure of its output distribution?
We may represent \DS\ as a small boolean circuit whose inputs are random bits and whose 18 output lines represent the diseases and symptom indicators.  
Specifically, assuming the parameters in Table~\ref{tab:probs} were dyadics, there would exist a circuit composed of constant-fan-in elements implementing \DS\ whose size grows linearly in the number of diseases and in the number of symptoms.

If we view the input lines as random variables, then the output lines of the logic gates are also random variables, and so we may ask: what conditional independencies hold among the circuit elements?
It is straightforward to show that the circuit diagram, viewed as a directed acyclic graph, is a directed graphical model capturing conditional independencies among the inputs, outputs, and internal gates of the circuit implementing \DS.
For every gate, the conditional probability mass function 
is characterized by the (constant-size) truth table of the logical gate. 

Therefore, if an efficient prior program samples from some distribution over a collection of binary random variables, then  those random variables exhibit many conditional independencies, in the sense that we can introduce a polynomial number of additional boolean random variables (representing intermediate computations) such that there exists a constant-fan-in directed graphical model over all the variables with constant-size conditional probability mass functions.

In Section~\ref{sec:latent} we return to the question of whether $\DS$ is a good model.
Here we conclude with a brief discussion of the history of graphical models in AI.

\subsection{Graphical models and AI}

Graphical models, and, in particular, directed graphical models or Bayesian networks, played a critical role in popularizing probabilistic techniques within AI in the late 1980s and early 1990s.  
Two developments were central to this shift: 
First, researchers introduced compact, computer-readable representations of distributions on large (but still finite) collections of random variables, and did so by explicitly representing a graph capturing conditional independencies and exploiting the factorization \eqref{gmfactor}.
Second, researchers introduced efficient graph-based algorithms that operated on these representations, exploiting the factorization to compute conditional probabilities.  
For the first time, a large class of distributions were given a formal representation that enabled the design of general purpose algorithms to compute useful quantities. 
As a result, the graphical model formalism became a lingua franca between practitioners designing large probabilistic systems, and figures depicting graphical models were commonly used to quickly communicate the essential structure of complex, but structured, distributions.

While there are sophisticated uses of Bayesian networks in cognitive science (see, e.g., \cite[\S3]{GKT08}), many models are not usefully represented by a Bayesian network.  In practice, this often happens when the number of variables or edges is extremely large (or infinite), but there still exists special structure that an algorithm can exploit to perform probabilistic inference efficiently.
In the next three sections, we will see examples of models that are not usefully represented by Bayesian networks, but which have concise descriptions as prior programs.

\section{Hierarchical models and learning probabilities from data}\label{sec:latent}

The \DS\ program makes a number of implicit assumptions that would deserve scrutiny in a real medical diagnosis setting.  For example, \DS\ models the diseases as \emph{a priori} independent, but of course, diseases often arise in clusters, e.g., as the result of an auto-immune condition.  In fact, because of the independence and the small marginal probability of each disease, there is an \emph{a priori} bias towards mutually exclusive diagnoses as we saw in the ``explaining away'' effect in \eqref{explaining}. The conditional independence of symptoms given diseases reflects an underlying casual interpretation of \DS\ in terms of diseases \emph{causing} symptoms.  In many cases, e.g., a fever or a sore neck, this may be reasonable, while in others, e.g., insulin resistance, it may not. 

Real systems that support medical diagnosis must relax the strong assumptions we have made in the simple \DS\ model, while at the same time maintaining enough structure to admit a concise representation.  In this and the next section, we show how both the structure and parameters in prior programs like \DS\ \emph{can be learned from data},
providing a clue as to how a mechanical mind could build predictive models of the world simply by experiencing and reacting to it.

\subsection{Learning as probabilistic inference}
\label{sec:parameterlearning}

The 95 probabilities in Table~\ref{tab:probs} eventually parameterize a distribution over $262\,144$
outcomes.  But whence come these 95 numbers?  As one might expect by studying the table of numbers, they were designed by hand to elucidate some phenomena and be vaguely plausible.  In practice, these parameters would themselves be subject to a great deal of uncertainty, and one might hope to use data from actual diagnostic situations to learn appropriate values.

There are many schools of thought on how to tackle this problem, but a hierarchical Bayesian approach
provides a particularly elegant solution 
that fits entirely within the \query\ framework.  The solution is to generalize the \DS\ program in two ways.  First, rather than generating one individual's diseases and symptoms, the program will generate data for $n+1$ individuals.  Second, rather than using the fixed table of probability values, the program will start by randomly generating a table of probability values, each independent and distributed uniformly at random in the unit interval, and then proceed along the same lines as \DS.  Let $\DS'$ stand for this generalized program.

The second generalization may sound quite surprising, and unlikely to work very well. The key is to consider the combination of the two generalizations.  To complete the picture, consider a past record of $n$ individuals and their diagnosis, represented as a (potentially partial) setting of the 18 variables $\{D_1,\dotsc,D_{11};\,S_1,\dotsc,S_7\}$.  We define a new predicate $\OS'$ that accepts the $n+1$ diagnoses generated by the generalized prior program $\DS'$ if and only if the first $n$ agree with the historical records, and the symptoms associated with the $n+1$'st agree with the current patient's symptoms.

What are typical outputs from $\query(\DS',\OS')$?  For very small values of $n$, we would not expect particularly sensible predictions, as there are many tables of probabilities that could conceivably lead to acceptance by $\OS'$.  However, as $n$ grows, some tables are much more likely to lead to acceptance.  In particular, for large $n$, we would expect the hypothesized marginal probability of a disease to be relatively close to the observed frequency of the disease, for otherwise, the probability of acceptance would drop.  This effect grows exponentially in $n$, and so we would expect that the typical accepted sample would quickly correspond with a latent table of probabilities that match the historical record.

\begin{figure}
\centering
\includegraphics[width=.5\linewidth]{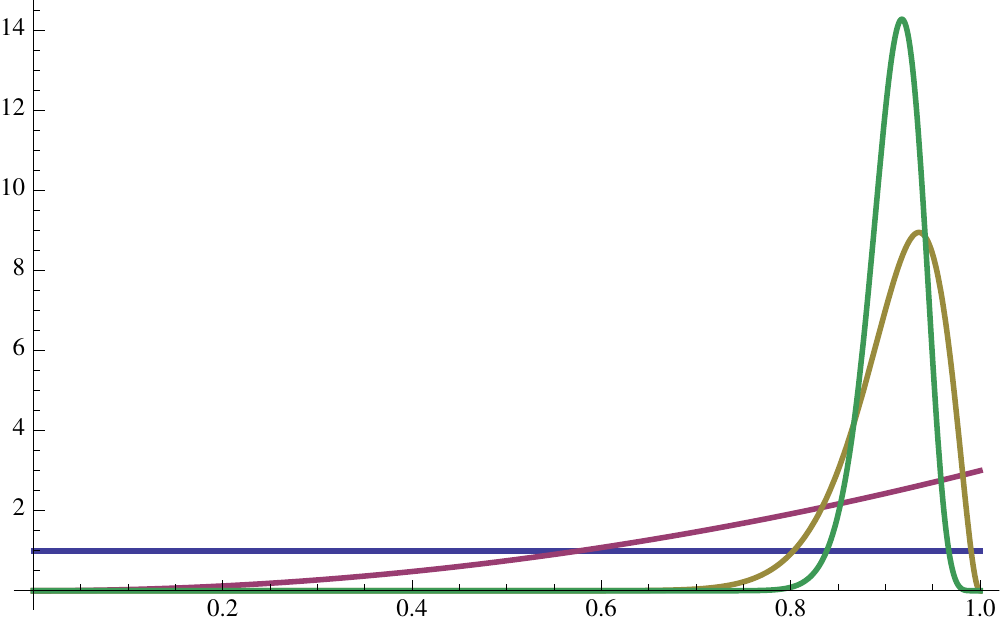}
\caption{Plots of the probability density of $\BetaD(a_1,a_0)$ distributions with density
$f(x; \alpha_1,\alpha_0) = \frac{\Gamma(\alpha_1+\alpha_0)}{\Gamma(\alpha_1)\Gamma(\alpha_0)} x^{\alpha_1-1}\,(1-x)^{\alpha_0-1}$
for parameters $(1,1)$, $(3,1)$, $(30,3)$, and $(90,9)$ (respectively, in height).  For parameters $\alpha_1,\alpha_0 > 1$, the distribution is unimodal with mean $\alpha_1/(\alpha_1 + \alpha_0)$.}
\label{fig:beta}
\end{figure}

We can, in fact, work out the conditional distributions of entries in the table in light of the $n$ historical records.  First consider a disease $j$ whose marginal probability, $p_j$, is modeled as a random variable sampled uniformly at random from the unit interval.  The likelihood that the $n$ sampled values of $D_j$ match the historical record is
\begin{align}\label{conddensity}
p_j^{k} \cdot (1-p_j)^{n-k},
\end{align}
where $k$ stands for the number of records where disease $j$ is present.  By Bayes' theorem, in the special case of a uniform prior distribution on $p_j$, the density of the conditional distribution of $p_j$ given the historical evidence is proportional to the likelihood \eqref{conddensity}.  This implies that, conditionally on the historical record, $p_j$ has a so-called $\BetaD(\alpha_1,\alpha_0)$ distribution with mean 
\begin{align*}
\frac{\alpha_1}{\alpha_1+\alpha_0} =
\frac{k+1}{n+2} \qquad
\end{align*}
and concentration parameter $\alpha_1+\alpha_0 = n+2$.
Figure~\ref{fig:beta} illustrates beta distributions under varying parameterizations, highlighting the fact that, as the concentration grows, the distribution begins to concentrate rapidly around its mean.  As $n$ grows, predictions made by $\query(\DS',\OS')$ will likely be those of runs where each disease marginals $p_j$ falls near the observed frequency of the $j$th disease.  In effect, the historical record data \emph{determines} the values of the marginals $p_j$.

A similar analysis can be made of the dynamics of the posterior distribution of the latent parameters $\ell_m$ and $c_{n,m}$, although this will take us too far beyond the scope of the present article.  Abstractly speaking, in finite dimensional Bayesian models like this one satisfying certain regularity conditions, it is possible to show that the predictions of the model converge to those made by the best possible approximation within the model to the distribution of the data.  (For a discussion of these issues, see, e.g., \cite{Bar98}.)

While the original \DS\ program makes the same inferences in each case, $\DS'$ learns to behave from experience.  The key to this power was the introduction of the latent table of probabilities, modeled as random variables.  This type of model is referred to as a \emph{hierarchical Bayesian model}.  The term ``Bayesian'' refers to the fact that we have modeled our uncertainty about the unknown probabilities by making them random and specifying a distribution that reflects our subjective uncertainty, rather than a frequency in a large random sample of patients.  The term ``hierarchy'' refers to the fact that in the graphical model representing the program, there is yet another level of random variables (the table of probabilities) sitting above the rest of the original graphical model.  More complicated models may have many more layers, or potentially even an infinite number of layers.

An interesting observation is that $\DS'$ is even more compact than $\DS$, as the specification of the distribution of the random table is logarithmic in the size of the table.  On the other hand, $\DS'$ relies on data to help it reduce its substantial \emph{a priori} uncertainty regarding these values.  This tradeoff---between, 
on the one hand,
the flexibility and complexity of a model and, on the other, the amount of data required in order to make sensible predictions---is seen throughout statistical modeling.  We will return to this point in Section~\ref{subsec:aspects}.

Here we have seen how the parameters
in prior programs can be modeled as random, and
thereby learned from data by conditioning on historical diagnoses.
In the next section, we consider the problem of learning not only the parameterization but the structure of the model's conditional independence itself.

\section{Random structure}
\label{sec:definetti}

Irrespective of how much historical data we have, $\DS'$ cannot go beyond the conditional independence assumptions implicit in the structure of the prior program.   
Just as we framed the problem of learning the table of probabilities as a probabilistic inference over a random table, 
we can frame the problem of identifying the correct structure of the dependence between symptoms and disease as one of probabilistic inference over random conditional independence \emph{structure} among the model variables.

In Section~\ref{sec:bayesnets}, we saw that conditional independence relationships among a collection of random variables can be captured by a directed acyclic graph.  
The approach we will discuss involves treating this graph as a random variable, whose distribution reflects our uncertainty about the statistical dependence among the diseases and symptoms before seeing data, and whose posterior distribution reflects our updated uncertainty about these relationships once the graph is forced to explain any evidence of dependence or independence in the historical data.

The model that we describe in this section introduces several additional layers and many more latent variables.
Outside of the Bayesian framework, these latent variables would typically be additional parameters that one
would tune to fit the data.
Typically, when one adds more parameters to a model, this improves the fit to the data at hand, but introduces a risk of ``overfitting'', which leads to poor predictive performance on unseen data.
 However, as we will see in Section~\ref{subsec:aspects}, the problem of overfitting is mitigated in this
Bayesian approach, because the latent variables are not optimized, but rather sampled conditionally.

\subsection{Learning structure as probabilistic inference }

Within AI and machine learning, the problem of learning a probabilistic model from data is a quintessential example of \emph{unsupervised learning}, 
and the approach of identifying a graph capturing conditional independence relationships among model variables is known as \emph{structure learning}.

In Section~\ref{sec:bayesnets} we saw that every distribution on $n$ binary random variables $X_1,\dotsc,X_n$  can be expressed in the form 
\begin{align}\label{gmfactorsec6}
p(X_1,\dotsc,X_k) = \prod_{j=1}^k p_j(X_j \given \PA(X_j) ).
\end{align}
where $G$ is a directed acyclic graph over the set $\{X_1,\dotsc,X_k\}$ of model variables; 
$\PA(X_j)$ denotes the parent vertices of $X_j$; 
and the $p_j(\cdot \given \cdot)$ are conditional probability mass functions specifying the distribution of each variable in terms of its parents' values.

From the perspective of this factorization, the tack we took in Section~\ref{sec:parameterlearning} was to assume that we knew the graphical structure $G$ (given by \DS) and learn (the parameterization of) the conditional mass functions by modeling them as random variables.  
We will now consider learning both ingredients simultaneously, and later pause to critique this strategy.

\subsection{A random probability distribution}

Let us return to the setting of medical diagnosis, and in particular the problem of modeling the presence/absence of the 11 diseases and 7 symptoms, represented by the variables $\{D_1,\dotsc,D_{11};\,S_1,\dotsc,S_7\}$.  

\newcommand{\RPD}{\newprogram{RPD}}
Towards this end, and with the factorization \eqref{gmfactorsec6} in mind, consider a prior program, which we will call $\RPD$ (for \emph{Random Probability Distribution}), that takes as input two positive integers $n$ and $D$ and produces as output $n$ independent samples from a random probability distribution on $\{0,1\}^D$.

Intuitively, \RPD\ works in the following way:  
First, \RPD\ generates a random directed acyclic graph $G$ with $D$ vertices.  Next, it generates a \emph{random} probability mass function $p$, which will specify a distribution over $D$ random variables, $X_1,\dotsc,X_D$.  
The probability mass function will be generated so that it satisfies the conditional independencies implied by the graph $G$ when it is viewed as a directed graphical model.   
The probability mass function $p$ is generated by choosing random conditional probability mass functions $p(X_j|\PA(X_j))$, one for each variable $X_j$ as in the factorization \eqref{gmfactorsec6}.  
Specifically,
if a variable $X_j$ has $k$ parents $\PA(X_j)$ (which collectively can take on $2^k$ possible $\{0,1\}$-assignments), then we must generate $2^k$ probabilities, one for each $\{0,1\}$-assignment $v$ of the parents, indicating the probability $p_{j|v}$ that $X_j = 1$ given that $\PA(X_j)=v$.  In particular, $p_j(1|v) = p_{j|v}$.
This fully determines $p$.  \RPD\ then proceeds to generate $n$ samples from $p$, each a list of $D$ binary values with the same distributions as $X_1,\dotsc,X_D$. 

More formally, \RPD\ begins by sampling a directed acyclic graph $G$ uniformly at random from the set $\mathcal G_D$ of all directed acyclic graphs over the vertex set $\{X_1,\dotsc,X_D\}$.  
For every vertex $j$ and every $\{0,1\}$-assignment $v$ to $X_i$'s parents $\PA(X_j)$, we sample a probability value $p_{j|v}$ uniformly at random from $[0,1]$.  
Let $j_1,\dotsc,j_D$ be a topological ordering of the vertices of $G$.  
We then repeat the following procedure $n$ times:
First, sample $X_{j_1} \in \{0,1\}$ with mean $p_{j_1 | ()}$, and then for $i=2,\dotsc,D$, sample $X_{j_i} \in \{0,1\}$ with mean $p_{j_i | v}$ where $v=(X_p : p \in \PA(j_i))$ is the $\{0,1\}$-assignment of $X_j$'s parents.  We then output the variables in order $X_1,\dotsc,X_D$, and repeat until we have produced $n$ such samples as output.

With \RPD\ fully specified, let us now consider the output of
\begin{align}\label{rpdoutput}
\query(\RPD(n+1,18), \OS')
\end{align}
where $\OS'$ is defined as in Section~\ref{sec:parameterlearning}, accepting $n+1$ diagnoses if and only if the first $n$ agree with historical records, and the symptoms associated with the $n+1$'st agree with the current patient's symptoms.  (Note that we are identifying each output $(X_1,\dotsc,X_{11},X_{12},\dotsc,X_{18})$ with a diagnosis $(D_1,\dotsc,D_{11},S_1,\dotsc,S_7)$, and have done so in order to highlight the generality of \RPD.)

As a first step in understanding \RPD, one can show that, conditioned on the graph $G$, the conditional independence structure of each of its $n$ outputs $(X_1,\dotsc,X_D)$ is precisely captured by $G$, when viewed as a Bayesian network (i.e., the distribution of the $X$'s satisfies the factorization~\eqref{gmfactorsec6}).  
It is then not hard to see that the probabilities $p_{j|v}$ parameterize the conditional probability mass functions, in the sense that $p(X_j = 1\given \PA(X_j) = v) = p_{j|v}$. Our goal over the remainder of the section will be to elucidate the posterior distribution on the graph and its parameterization, in light of historical data.

\newcommand{\njv}{n_{j|v}}
\newcommand{\kjv}{k_{j|v}}
To begin, we assume that we know the graph $G$ for a particular output from \eqref{rpdoutput}, and then study the likely values of the probabilities $p_{j|v}$ conditioned on the graph $G$.  Given the simple uniform prior distributions, we can in fact derive analytical expressions for the posterior distribution of the probabilities $p_{j|v}$ directly, conditioned on historical data and the particular graph structure $G$.  In much the same was as our analysis in Section~\ref{sec:parameterlearning}, it is easy to show that the expected value of $p_{j|v}$ on those runs accepted by \query\  is
\begin{align*}
\frac {\kjv + 1}{\njv + 2}
\end{align*}
where $\njv$ is the number of times in the historical data where the pattern $\PA(X_j) = v$ arises; and $\kjv$ is the number of times when, moreover, $X_j = 1$.  This is simply the ``smoothed'' empirical frequency of the event $X_j =1$ given $\PA(X_j) = v$.  In fact, the $p_{j|v}$ are conditionally Beta distributed with concentration $\njv+2$.
Under an assumption that the historical data are conditionally independent and identically distributed according to a measure $P$, it follows by a law of large numbers argument that these probabilities converge almost surely to the underlying conditional probability $P(X_j =1 | \PA(X_j) = v )$ as $n \to \infty$.

The variance of these probabilities is one characterization of our uncertainty, and for each probability $p_{j|v}$, the variance is easily shown to scale as $\njv^{-1}$, i.e., the number of times in the historical data when $\PA(X_j) = v$.  
Informally, this suggests that, the smaller the parental sets (a property of $G$), the more certain we are likely to be regarding the correct parameterization, and, in terms of \query, the smaller the range of values of $p_{j|v}$ we will expect to see on accepted runs.  
This is our first glimpse at a subtle balance between the simplicity of the graph $G$ and how well it
captures hidden structure in the data.

\subsection{Aspects of the posterior distribution of the graphical structure}
\label{subsec:aspects}

The space of directed acyclic graphs on $18$ variables is enormous, and computational hardness results 
\cite{Coo90,DL93,CSH08}
imply there will be no simple way to summarize the structure of the posterior distribution, at least not one that suggests an efficient method in general for choosing structures with high posterior probability.  It also goes without saying that one should not expect the PTM defined by \eqref{rpdoutput} to halt 
within a reasonable time for any appreciable value of $n$
because the probability of generating the structure that fits the data
is astronomically small.
However it is still instructive to understand the conceptual structure of the posterior distribution of the graph $G$.  
On the one hand, there are algorithms that operate quite differently from the naive mechanism of \query\ and  work reasonably well in practice at approximating the task defined here, despite hardness results.
There are also more restricted, but still interesting, versions of this task for which there exist
algorithms that work remarkably well in practice and sometimes provably so
\cite{BJ03}.

On the other hand, this example is worth studying because it reveals an important aspect of some hierarchical Bayesian models with regard to their ability to avoid ``overfitting'', and gives some insight into why we might expect ``simpler'' explanations/theories to win out in the short term over more complex ones.

Consider the set of probability distributions of the form \eqref{gmfactorsec6} for a particular graph $G$. We will
refer to these simply as the \emph{models in $G$} when there is no risk of confusion.
The first observation to make is that if a graph $G$ is a strict subgraph of another graph $G'$ on the same vertex set, then the set of 
models in $G$ is a strict subset of those 
in $G'$.  It follows that, no matter the data set, the best-fitting probability distribution corresponding with $G'$ will be no worse than the best-fitting model 
in $G$.  Given this observation, one might guess that samples from \eqref{rpdoutput} would be more likely to come from models whose graphs have more edges, as such graphs always contain a model that fits the historical data better.

However, the truth is more subtle.  Another key observation is that the posterior probability of a particular graph $G$ does not reflect the best-fitting model in $G$, but rather reflects the \emph{average} ability of models in $G$ to explain the historical data.  In particular, this average is over the random parameterizations $p_{j|v}$ of the conditional probability mass functions.  Informally speaking, if a spurious edge exists in a graph $G'$, a typical distribution from $G'$ is less likely to explain the data than a typical distribution from the graph with that edge removed.

In order to characterize the posterior distribution of the graph, we can study the likelihood that a sample from the prior program is accepted, assuming that it begins by sampling a particular graph $G$.  We begin by focusing on the use of each particular probability $p_{j|v}$, and note that every time the pattern $\PA(X_j)=v$ arises in the historical data, the generative process produces the historical value $X_j$ with probability $p_{j|v}$ if $X_j=1$ and $1-p_{j|v}$ if $X_j=0$.  It follows that the probability that the generative process, having chosen graph $G$ and parameters $\{p_{j|v}\}$, proceeds to produce the historical data is
\begin{align}\label{eq:jp}
\prod_{j=1}^D \prod_{v} p_{j|v}^{\kjv} (1-p_{j|v})^{\njv - \kjv},
\end{align}
where $v$ ranges over the possible $\{0,1\}$ assignments to $\PA(X_j)$ and $\kjv$ and $\njv$ are defined as above.  In order to determine the probability that the generative process produces the historical data (and thus is accepted), assuming only that it has chosen graph $G$, we must take the expected value of \eqref{eq:jp} with respect to the uniform probability measure on the parameters, i.e., we must calculate the marginal probability of the historical data conditioned the graph $G$.  Given the independence of the parameters, it is straightforward to show that this expectation is
\begin{align}\label{score}
\score(G) \defas 
\prod_{j=1}^D \prod_{v}
(\njv+1)^{-1} \binom{\njv}{\kjv}^{-1}
\end{align}
Because the graph $G$ was chosen uniformly at random, it follows that the posterior probability of a particular graph $G$ is proportional to $\score(G)$.  

We can study the preference for one graph $G$ over another $G'$ by studying the ratio of their scores:
\begin{align*}
\frac{\score(G)}{\score(G')}\ .
\end{align*}
This score ratio is known as the \emph{Bayes factor}, which Good termed the
\emph{Bayes--Jeffreys--Turing factor} \cite{Goo68,Goo75}, and which Turing
himself called the \emph{factor in favor of a hypothesis}
(see \cite{Goo68}, \cite[\S1.4]{Zab12}, and \cite{Tur12}).
Its logarithm is
sometimes known as the \emph{weight of evidence} \cite{Goo68}.
The form of \eqref{score}, a product over the local structure of the graph, reveals that the Bayes factor will depend only on those parts of the graphs $G$ and $G'$ that differ from each other. 

\begin{figure}
\centering
\includegraphics[width=.45\linewidth]{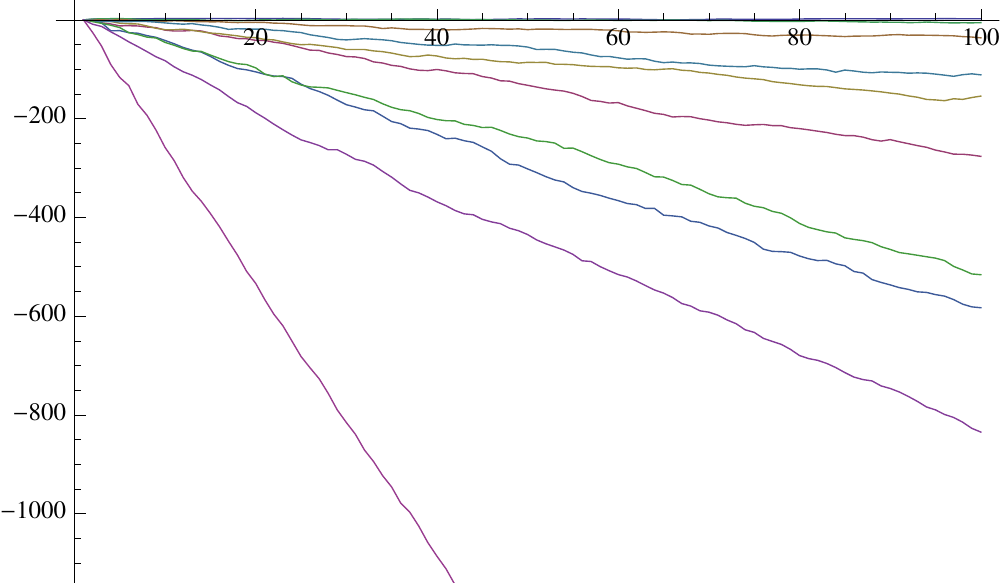}
\includegraphics[width=.45\linewidth]{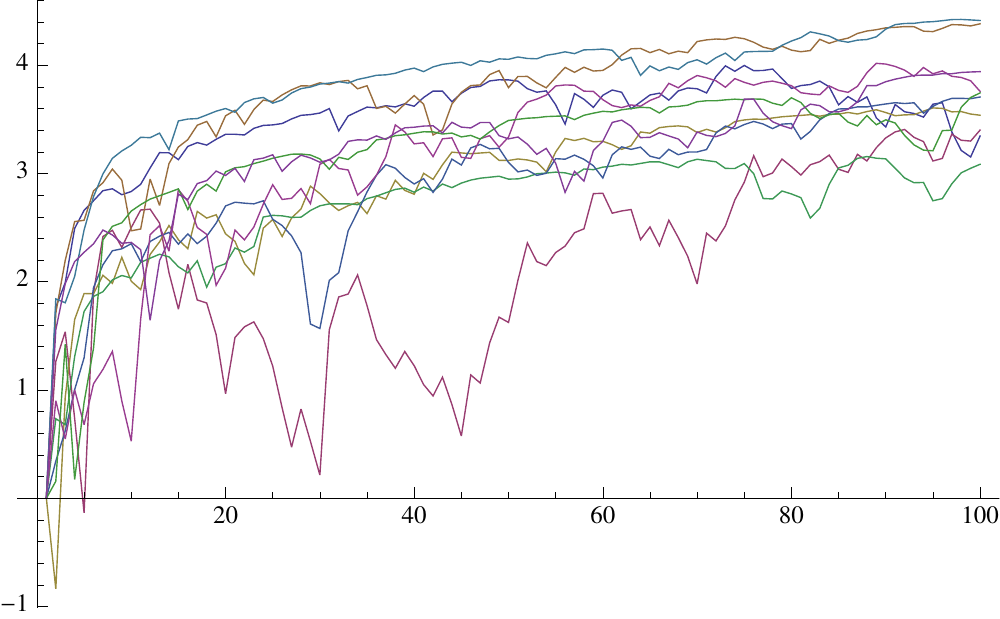}

\caption{Weight of evidence for independence versus dependence (positive values support independence) of a sequence of pairs of random variables sampled from $\RPD(n,2)$.  
(left) 
When presented with data from a distribution where $(X,Y)$ are indeed dependent, the weight of evidence rapidly accumulates for the dependent model, at an asymptotically linear rate in the amount of data.
(right) 
When presented with data from a distribution where $(X,Y)$ are independent, the weight of evidence slowly accumulates for the independent model, at an asymptotic rate that is logarithmic in the amount of data.
Note that the dependent model can imitate the independent model, but, on average over random parameterizations of the conditional probability mass functions, the dependent model is worse at modeling independent data.
}
\label{fig:structure}
\end{figure}

\newcommand{\njvY}{n}
\newcommand{\kjvY}{k}
\newcommand{\njvYi}{n_1}
\newcommand{\kjvYi}{k_1}
\newcommand{\njvYo}{n_0}
\newcommand{\kjvYo}{k_0}
Consider the following simplified scenario, which captures several features of learning structure from data:  Fix two graphs, $G$ and $G'$, over the same collection of random variables, but assume that in $G$, two of these random variables, $X$ and $Y$, have no parents and are thus independent, and in $G'$ there is an edge from $X$ to $Y$, and so they are almost surely dependent.  From \eqref{score}, we may deduce that the score ratio is
\begin{align}\label{gscore}
\frac
  {(\njvYi+1) (\njvYo+1)} {(\njvY+1)} 
\frac 
  {\binom{\njvYi}{\kjvYi}  \binom{\njvYo}{\kjvYo} }
  {\binom{\njvY}{\kjvY}},
\end{align}
where 
$\njvY$ counts the total number of observations; 
$\kjvY$ counts $Y=1$; 
$\njvYi$ counts $X=1$; 
$\kjvYi$ counts $X=1$ and $Y=1$; 
$\njvYo$ counts $X=0$; and
$\kjvYo$ counts $X=0$ and $Y=1$.
In order to understand how the Bayes factor \eqref{gscore} for graph $G$ behaves, let us first consider the case where $G'$ is the true underlying graph, i.e., when $Y$ is indeed dependent on $X$.  Using the law of large numbers, and Stirling's approximation, we can reason that the evidence for $G'$ accumulates rapidly, satisfying
\begin{align*}
\log \frac{\score(G)}{\score(G')} \sim - C \cdot \njvY, \quad\as,
\end{align*}
for some constant $C>0$ that depends only on the joint distribution of $X$ and $Y$.  As a concrete example, when $X$ and $Y$ have mean $\frac 1 2$, the constant is given by
\begin{align*}
\log \frac {(1-d)^{d-\frac12}}{(1+d)^{d+\frac12}},
\end{align*}
where $d = \Pr\{Y=1 | X=1\} = 1- \Pr\{Y=1 | X=0 \}$.  For example, $C \to 0$ as $d\downarrow0$; $C \approx 0.13$ when $d=1/2$; and $C$ achieves its maximum, $\log 2$, as $d \uparrow 1$.
The first plot in Figure~\ref{fig:structure} shows the progression of the weight of evidence when data is drawn from distributions generated uniformly at random to satisfy the conditional independencies captured by $G'$.  As predicted, the evidence rapidly accumulates at a linear rate in favor of $G'$. 

On the other hand, when $G$ is the true underlying graph and $Y$ is independent and $X$, 
one can show using similar techniques to above that
\begin{align*}
\log \frac{\score(G)}{\score(G')} \sim \frac 1 2 \log \njvY, \quad\as
\end{align*}
The second plot in Figure~\ref{fig:structure} shows the progression of the weight of evidence when data is drawn from distributions generated uniformly at random to satisfy the conditional independencies captured by $G$.  As predicted, the evidence accumulates, but at a much slower logarithmic rate. 

In both cases, evidence accumulates for the correct model.  In fact, it can be shown that the expected weight of evidence is always non-negative for the true hypothesis, a result due to Turing himself \cite[p.~93]{Goo91}.  
Because the prior probabilities for each graph are fixed and do not vary with the amount of data, the weight of evidence will  eventually eclipse any prior information and determine the posterior probability.  On the other hand, as we have seen, the evidence accumulates rapidly for dependence and much more slowly for independence and so we might choose our prior distribution to reflect this imbalance, preferring graphs with fewer edges \emph{a priori}.\footnote{This analysis in terms of Bayes factors also aligns well with experimental findings on human judgments of evidence for causal structure (see, e.g., \cite{GT05}).}

\subsection{Bayes' Occam's razor}
In the example above when $X$ and $Y$ are independent, we see that evidence accumulates for the simpler graph over the more complex graph, despite the fact that there is almost always a parameterization of the more complex graph that assigns a higher likelihood to the data than any parametrization of the simpler graph. 
This phenomenon is known as Bayes' Occam's razor \cite[Ch.~28]{Mac03}, and it represents a natural way in which hierarchical models like \RPD\ with several layers---a random graph, random conditional probability mass functions generated given the graph, and finally, the random data generated given the graph and the conditional probability mass functions---end up choosing models with intermediate complexity.

One way to understand this phenomenon is to note that, if a model has many degrees of freedom, then each configuration must be assigned, on average, less probability than it would under a simpler model with fewer degrees of freedom.  Here, a graph with additional edges has more degrees of freedom, and while it can represent a strictly larger set of distributions than a simpler graph, a distribution with simple graphical structure $G$ is assigned greater probability density under $G$ than under a more complex graph. 
Thus, if fewer degrees of freedom suffice, the simpler model is preferred.

We can apply this same perspective to $\DS$, $\DS'$ and \RPD:
The \RPD\ model has many more degrees of freedom than both \DS\ and $\DS'$.  In particular, given enough data, \RPD\ can fit any distribution on a finite collection of binary variables, as opposed to $\DS'$, which cannot because it makes strong and immutable assumptions.  On the other hand, with only a small amount of training data, one would expect the \RPD\ model to have high posterior uncertainty.
Indeed, one would expect much better predictions from $\DS'$ versus $\RPD$, if both were fed data generated by $\DS$, especially in the low-data regime.  

An important research problem is bridging the gap between \RPD\ and $\DS'$.
Whereas $\DS'$ makes an immutable choice for one particular structure,  
\RPD\ assumes \emph{a priori} that every graphical structure is equally likely to explain the data.
If, instead, one were uncertain about the
structure but expected to find some particular regular pattern in the graphical structure, one could define an alternative model $\RPD'$ that placed a non-uniform distribution on the graph, favoring such patterns, and one could then expect better predictions when that pattern was indeed present in the data. 
However, one often does not know exactly which pattern might arise.  But in this case, we can take the same step we took when defining \RPD\ and consider a \emph{random} pattern, drawn from some space of possible patterns.
This would constitute an additional level to the hierarchical model.
Examples of this idea are described by Mansinghka et al.\ \cite{MKTG06} and  
Kemp et al.~\cite{KSBT07}, and this technique constitutes one aspect of the
general approach of ``theory-based Bayesian models'' \cite{GT06,TGK06, GKT08, KT08, GT09}.

Up until this point, we have considered the problem of reasoning and representing our own uncertainty in light of evidence.  However, in practice, representations of uncertainty are often useful because they support decision making under uncertainty.  In the next section, we show how the \query\ framework can be used to turn models of our uncertainty, including models of the effects of our own actions, into decisions.

\section{Making decisions under uncertainty}
\label{sec:tom}

Until now, we have discussed how computational processes can represent uncertain knowledge, and how these processes can be transformed using \query\ to reflect our uncertainty after incorporating new evidence.  
In this section, we consider the problem of making decisions under uncertainty, which will require us to reason not only about the immediate effects of the actions we take, but also about future decisions and the uncertainty we are likely to face when making them.  In the end, we will give a recursive characterization of an approximately optimal action, and show how this relates to simple feedback rules that Turing himself proposed.

The solution we describe models decisions as random variables and decision making as sampling.  
Using PTMs and \query, we construct distributions over actions a decision maker might take after reasoning about the effects of those actions and their likely success in achieving some goal or objective.
The particular distributions we will construct are based in part on the
exponentiated choice rule introduced by Luce~\cite{Luc59, Luc77} in the context of modeling human choices.

Our presentation extends that for the ``fully observable'' case given by
Goodman, Mansinghka, Roy, Bonawitz, and Tenenbaum \cite{GMRBT08} and Mansinghka \cite[\S2.2.3]{Man09}.
In particular, recursion plays a fundamental role in our solution, and thus pushes us beyond the representational capacity of many formalisms for expressing complex probabilistic models. 
Even more so than earlier sections, the computational processes we define with \query\ will not be serious proposals for \emph{algorithms}, although they will define distributions for which we might seek to implement approximate inference. However, those familiar with traditional presentations may be surprised by the ease with which we move between problems often tackled by distinct formalisms and indeed, this is a common feature of the \query\ perspective.

\subsection{Diagnosis and Treatment}

Returning to our medical diagnosis theme, consider a doctor faced with choosing between one or more treatment plans.  
What recommendation should they give to the patient and how might we build a system to make similar choices?

In particular, imagine a patient in need of an organ transplant and the question of whether to wait for a human donor or use a synthetic organ.  
There are a number of sources of uncertainty to contend with:  
While waiting for a human donor, the patient may become too ill for surgery, risking death.  
On the other hand, the time before a human organ becomes available would itself be subject to uncertainty.  
There is also uncertainty involved post-operation: 
will the organ be accepted by the patient's immune system without complication? 
How long should the patient expect to live in both conditions, taking into consideration the deterioration that one should expect if the patient waits quite a while before undergoing a transplant?

This situation is quite a bit more complicated.  
The decision as to whether to wait changes daily as new evidence accumulates, and how one should act today depends implicitly on the possible states of uncertainty one might face in the future and the decisions one would take in those situations.  
As we will see, we can use \query, along with models of our uncertainty, to make decisions in such complex situations.

\subsection{A single decision}

\newcommand{\RT}{\newprogram{RT}}
\newcommand{\SIM}{\newprogram{SIM}}
\newcommand{\CT}{\newprogram{CT}^*}
\newcommand{\CTADD}{\newprogram{CT}^+}
\newcommand{\xx}{x}
\newcommand{\yy}{y}

We begin with the problem of making a single decision between two alternative treatments.  
What we have at our disposal are two simulations $\SIM_\xx$ and $\SIM_\yy$ capturing our uncertainty as to the effects of those treatments.  
These could be arbitrarily complicated computational processes, but in the end we will expect them to produce an output 1 indicating that the resulting simulation of treatment was successful/acceptable and a 0 otherwise.\footnote{Our discussion is couched in terms of successful/unsuccessful outcomes,
rather than in terms of a real-valued loss (as is standard in classical decision theory).  However, it is possible to move between these two formalisms with additional hypotheses, e.g., boundedness and continuity of the loss.  See \cite{THS06} for one such approach.}
In this way, both simulations act like predicates, and so we will say that a simulation accepts when it outputs 1, and rejects otherwise.  
In this section we demonstrate the use of PTMs and \query\ to define distributions over \emph{actions}---here, treatments---that are likely to lead to successful outcomes.   

Let $\RT$ (for \emph{Random Treatment}) be the program that chooses a treatment $Z\in\{\xx,\yy\}$ uniformly at random and consider the output of the program
\begin{align*}
\CT = \query(\RT,\SIM_Z),
\end{align*}
where $\SIM_Z$ is interpreted as the program that simulates $\SIM_z$ when $\RT$ outputs $Z=z \in \{\xx,\yy\}$.  The output of $\CT$ is a treatment, $\xx$ or $\yy$, and we will proceed by interpreting this output as the treatment chosen by some decision maker.

With a view towards analyzing $\CT$ (named for \emph{Choose Treatment}), 
let $p_z$ be the probability that $\SIM_z$ accepts (i.e., $p_z$ is the probability that treatment $z$ succeeds), and assume that $p_\xx > p_\yy$.  Then $\CT$ ``chooses'' treatment $\xx$ with probability
\begin{align*}
\frac{p_\xx}{p_\xx + p_\yy}
= \frac{\rho}{\rho+1}
\end{align*}
where $\rho \defas p_\xx/p_\yy$ and $\rho > 1$ by assumption.  It follows that $\CT$ is more likely to choose treatment $\xx$ and that the strength of this preference is controlled by the multiplicative ratio $\rho$ (hence the multiplication symbol $*$ in $\CT$). If $\rho \gg 1$, then  treatment $\xx$ is chosen essentially every time.  

\begin{figure}
\centering
\includegraphics[width=.4\linewidth]{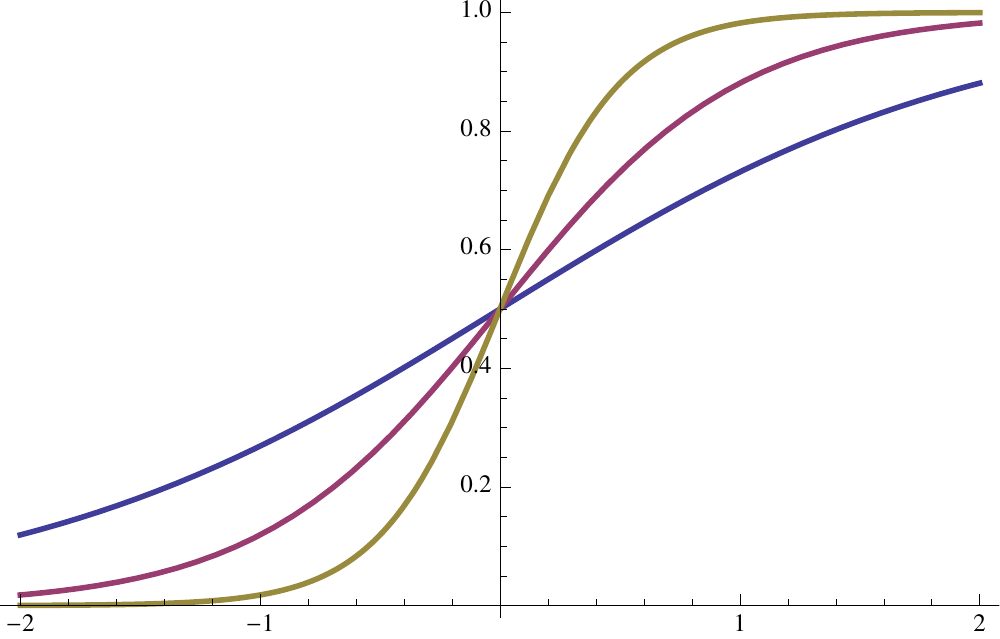}
\includegraphics[width=.4\linewidth]{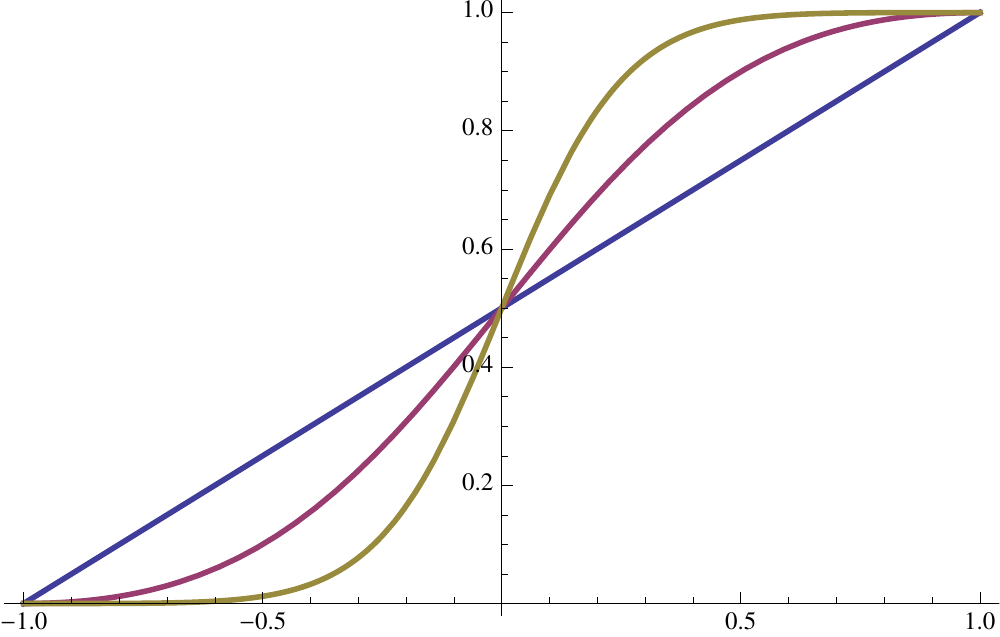}
\caption{Plots of the sigmoidal curves arising from the probability of treatment $\xx$ under $\CT_k$ (left) and $\CTADD_k$ (right) as a function of the log probability difference $\log \rho = \log p_\xx - \log p_\yy$ and probability difference $\alpha = p_\xx - p_\yy$, respectively, between the treatments. Here $k \in \{1,2,4\}$.  The straight line on the right corresponds to $k=1$, and curvature increases with $k$.}
\label{fig:CTADD}
\end{figure}

\newcommand{\REPEAT}{\newprogram{REPEAT}}
On the other hand, even if treatment $\xx$ is twice as likely to succeed, $\CT$ still chooses treatment $\yy$ with probability $1/3$.
One might think that a person making this decision should always choose treatment $\xx$.  However, it should not be surprising that $\CT$ does not represent this behavior because it accepts a proposed action solely on the basis of a single successful simulation of its outcome. 
With that in mind, let $k \ge 0$ be an integer, and consider the program
\begin{align*}
\CT_k = \query(\RT, \REPEAT(k,\SIM_Z)),
\end{align*}
where the machine \REPEAT\ on input $k$ and $\SIM_Z$ accepts if $k$ independent simulations of $\SIM_Z$ all accept. 
The probability that $\CT_k$ chooses treatment $\xx$ is 
\begin{align*}
\frac {\rho^k} {\rho^k+1}
\end{align*}
and so a small multiplicative difference between $p_\xx$ and $p_\yy$ is exponentiated, and $\CT_k$  chooses the more successful treatment with all but vanishing probability as $k\to \infty$. (See the left plot in Figure~\ref{fig:CTADD}.) Indeed, in the limit, $\CT_\infty$ would always choose treatment $\xx$ as we assumed $p_\xx > p_\yy$.
(For every $k$, this is the well-known exponentiated Luce choice rule \cite{Luc59, Luc77}.)

The behavior of $\CT_\infty$ agrees with that of a classical decision-theoretic approach, where, roughly speaking, one fixes a loss function over possible outcomes and seeks the action minimizing the expected loss.  (See 
\cite{DeG05} for an excellent resource on statistical decision theory.)
On the other hand, classical decision theory, at least when applied naively, often fails to explain human
performance:  It is well-documented that human behavior does not agree with mathematically ``optimal''
behavior with respect to straightforward formalizations of decision problems that humans face. (See Camerer~\cite{Cam11} for a discussion of models of actual human performance in strategic situations.)

While we are not addressing the question of designing efficient algorithms, there is also evidence that
seeking optimal solutions leads to computational intractability.  (Indeed, PSPACE-hardness
\cite{PT87}
and even undecidability \cite{MHC03}
can arise in the general case of certain standard formulations.) 

Some have argued that human behavior is better understood in terms of a large collection of heuristics.  For example, Goldstein and Gigerenzer~\cite{GG02} propose the ``recognition heuristic'', which says that when two objects are presented to a human subject and only one is recognized, that the recognized object is deemed to be of greater intrinsic value to the task at hand.
The problem with such explanations of human behavior (and approaches to algorithms for AI) is that they often do not explain how these heuristic arise.  Indeed, a theory for how such heuristics are learned would be a more concise and explanatory description of the heuristics than the heuristics themselves.  A fruitful approach, and one that meshes with our presentation here, is to explain heuristic behavior as arising from approximations to rational behavior, perhaps necessitated by intrinsic computational hardness.  Human behavior would then give us clues as to which approximations are often successful in practice and likely to be useful 
for algorithms.

In a sense, $\CT_k$ could be such a model for approximately optimal behavior.  However,
since $\CT_k$ chooses an action on the basis of the ratio $\rho = p_\xx/p_\yy$, one problematic feature is its sensitivity to small
differences $|p_\xx - p_\yy |$ in the absolute probabilities of success when $p_\xx,p_\yy \ll 1$.
Clearly, for most decisions, a $1/10\,000$ likelihood
of success is not appreciably better than a $1/20\,000$ likelihood.
A result by Dagum, Karp, Luby and Ross~\cite{DKLR00} on estimating small probabilities to high relative accuracy suggests that this sensitivity in $\CT_k$ might be a potential source of computational hardness in efforts to design algorithms, not to mention a point of disagreement with human behavior.  It stands to reason that it may be worthwhile to seek models that do not exhibit this sensitivity.

\newcommand{\MAJ}{\newprogram{MAJ}}
\newcommand{\BER}{\newprogram{BERNOULLI}}
To this end, let $\SIM_\xx$ and $\SIM_\yy$ be independent simulations and
consider the predicate $\MAJ_\xx$ (named for \emph{Majority}) that accepts if $\SIM_\xx$ succeeds and $\SIM_\yy$ fails, rejects in the opposite situation, and chooses to accept or reject uniformly at random otherwise.
We then define
\begin{align*}
\CTADD = \query(\RT, \MAJ_Z),
\end{align*}
It is straightforward to show that treatment $\xx$ is chosen with probability
\begin{align*}
\frac {1 + (p_\xx-p_\yy)}2 = \frac {1 + \alpha}{2}
\end{align*}
and so the output of $\CTADD$ is sensitive to only the additive difference $\alpha = p_\xx-p_\yy$ (hence the addition symbol $+$ in $\CTADD$). In particular, when
$p_\xx \approx p_\yy$,  $\CTADD$ chooses an action nearly uniformly at random.   Unlike $\CT$, it is the case that $\CTADD$ is insensitive to the ratio $p_\xx/p_\yy$ when $p_\xx,p_\yy \approx 0$.

Similarly to $\CT_k$, we may define $\CTADD_k$ by
\begin{align*}
\CTADD_k = \query(\RT, \REPEAT(k, \MAJ_Z)),
\end{align*}
in which case it follows that $\CTADD_k$ accepts treatment $\xx$ with probability
\begin{align*}
\frac{1}{1+\bigl(\frac{1-\alpha}{1+\alpha}\bigr)^k}.
\end{align*}
Figure~\ref{fig:CTADD} shows how this varies as a function of $\alpha$ for several values of $k$.
Again, as $k \to \infty$, the decision concentrates on the treatment with the greatest probability of success, although there is always a region around $\alpha = 0$ where each treatment is almost equally likely to be chosen.  

In this section, we have shown how a model of one's uncertainty about the likely success of a single action can be used to produce a distribution over actions that concentrates on actions likely to achieve success.  
In the next section, we will see how the situation changes when we face multiple decisions.  
There, the likely success of an action depends on future actions, which in turn depend on the likely success of yet-more-distant actions.  
Using recursion, we can extend strategies for single decisions to multiple decisions,  defining distributions over sequences of actions that, under appropriate limits, agree with notions from classical decision theory, but also suggest notions of approximately optimal behavior.

\subsection{Sequences of decisions}

How can we make a sequence of good decisions over time, given a model of their effects?
Naively, we might proceed along the same lines as we did in the previous section, sampling now a \emph{sequence} of actions conditioned on, e.g., a simulation of these actions leading to a successful outcome.
Unfortunately, this does not lead to a sensible notion of approximately optimal behavior, as the first action is chosen on the basis of a fixed sequence of subsequent actions that do not adapt to new observations.  
Certainly one should react differently when a door leads not to the next room but to a closet!

In order to recover classical notions of optimality under an appropriate limit, we need to evaluate exhaustive plans---called \defn{policies} in the planning literature---that specify how to act in every conceivable situation that could arise.  Indeed, the optimal action to take at any moment is that which, when followed by optimal behavior thereafter, maximizes the probability of a successful outcome.  As one might expect, the  self-referential nature of optimal behavior will lead us to recursive definitions.

Returning to our transplant scenario, each day we are faced with several possible options: waiting another day for a human donor match; running further diagnostic tests; choosing to go with the synthetic organ; etc.
As time passes, observations affect our uncertainty.  Observations might include the result of a diagnostic test, the appearance of the patient, how they feel, news about potential donor matches, etc.
Underlying these observations (indeed, \emph{generating} these observations) are a network of processes: 
the dynamics of the patient's organ systems, the biophysical mechanisms underlying the diagnostic tests included in our observations, the sociodynamics of the national donor list, etc.
  
Observations are the channel through which we can make inferences about
the state of the underlying processes and, by reducing our uncertainty, make better decisions.  
On the other hand, our actions (or inaction) will influence the evolution of the underlying latent processes, and so, 
in order to choose good actions, we must reason not only about future observations (including eventual success or failure) but our own future actions.

\newcommand{\EXTEND}{\newprogram{NEXTSTATE}}
While their may be many details in any particular sequential decision task, we can abstract away nearly all of them.  In particular, at any point, the sequence of observations and actions that have transpired constitutes our \defn{belief state}, and our model of the underlying latent processes and the effects of our actions boils down to a description of how our belief state evolves as we take actions and make observations.  More concretely, a model for a sequential decision task is captured by a 
PTM \EXTEND, which takes a belief state and an action as input and returns the new, random belief state arising from making an additional observation.  Certain belief states are \emph{terminal} and correspond either with a successful or unsuccessful outcome.

The internal details of \EXTEND\ can be arbitrarily complex, potentially representing faithful attempts at simulating the types of processes listed above, e.g., employing detailed models of physiology, diagnostic techniques, the typical progression of the national donor list, success rates of organ transplants, life expectancy under various conditions, etc.  Our goal is to transform the computational process \EXTEND\ characterizing the sequential decision task into a computational process representing a \emph{stochastic} policy that maps belief states to \emph{distributions on} actions that have a high probability of success.

\newcommand{\OUTCOME}{\newprogram{OUTCOME}}
To begin, we describe a PTM \OUTCOME, which uniformly in a belief state $b$ and index $\pi$ for a stochastic policy, simulates an outcome resulting from following the policy $\pi$, starting from a belief state $b$.  We may describe the behavior of  \OUTCOME\ inductively as follows:  First, a check is made as to whether $b$ is a terminal belief state.  If so, the machine halts and returns 1 (accept) or 0 (reject) depending on whether $b$ represents a successful or unsuccessful outcome, respectively.  Otherwise, \OUTCOME\ evaluates $\pi(b)$, producing a random action $a$, and then performs an independent simulation of $\EXTEND(b,a)$ to produce a new belief state $b'$, at which point the process repeats anew.  
In order to simplify the analysis, we will make the following assumptions: First, we will assume that there is some positive integer $M$ such that, for every belief state $b$ and policy $\pi$, we have that $\OUTCOME(b,\pi)$ halts within $M$ iterations.  Second, for every non-terminal belief state $b$ and policy $\pi$, we will assume that $\OUTCOME(b,\pi)$ accepts with positive probability. 

\newcommand{\RA}{\newprogram{RA}}
\newcommand{\POLICY}{\newprogram{POLICY}}
\newcommand{\VALUE}{\newprogram{VALUE}}
\newcommand{\CAA}{\newprogram{ACT}}
We can now cast the problem of choosing the first of a sequence of actions into the single decision framework as follows:
Let $\SIM_{b,\pi,z}$ be the PTM that, uniformly in a belief state $b$, index $\pi$ for a stochastic policy, and action $z$, 
 simulates $\EXTEND(b,z)$, producing an updated belief state $b'$, and then simulates $\OUTCOME(b',\pi)$, randomly accepting or rejecting depending on whether the simulation of the policy $\pi$ starting from $b'$ resulted in a successful outcome or not.
Let \RA\ be a PTM that samples an action uniformly at random.
Then
\begin{align}\label{howtoact}
\CAA(b,\pi) \defas \query(\RA, \SIM_{b,\pi,Z})
\end{align}
represents a distribution on actions that concentrates more mass on the action leading to a higher probability of success under the policy $\pi$. Here $\SIM_{b,\pi,z}$ plays a role similar to that played by $\SIM_z$ in the single decision framework described earlier.  The two additional inputs are needed because we must assign 
an action (or more accurately, a distribution over actions)
to every belief state and policy.
As before, we can amplify our preference for the action having the higher probability of success by asking for $k$ simulations to succeed.

In order to determine a complete policy, we must specify the policy $\pi$ that governs future behavior.  
The key idea is to choose actions in the future according to~\eqref{howtoact} as well, and we can implement this idea using recursion. In particular, by Kleene's recursion theorem, there is a PTM \POLICY\ satisfying
\begin{align}\label{recdefn}
\POLICY(b) = \CAA(b,\POLICY).
\end{align}
The simplifying assumptions we made above are enough to guarantee that \POLICY\ halts with probability one on every belief state.  Those familiar with Bellman's ``principle of optimality'' \cite{Bel57} will notice that \eqref{howtoact} and \eqref{recdefn} are related to the value iteration algorithm for Markov decision processes \cite{How60}.

In order to understand the behavior of \POLICY, consider the following simple scenario: a patient may or may not have a particularly serious disease, but if they do, an special injection will save them.  On the other hand, if a patient is given the same injection but does not have the condition, there is a good chance of dying from the injection itself.  Luckily, there is a diagnostic test that reliably detects the condition.  How would \POLICY\ behave?  

\begin{figure}[t]
\begin{tikzpicture}[grow=right,font=\footnotesize,every tree node/.style={draw=none,rectangle},sibling
distance=14pt, level distance=2.5cm]
\tikzset{edge from parent/.style={draw,->,sloped,edge from parent path=
    {(\tikzparentnode) -> (\tikzchildnode)}}}
\tikzset{Action/.style =   {draw,font=\footnotesize,fill=white}}
\tikzset{Pr/.style =   {draw=none,fill = white}}    
\Tree [.$\varepsilon$ 
            \edge node[Action] {WAIT};
            [.$\square$  
                          \edge node[Pr] {$0.45$}; 
                          Dead
                          \edge node[Pr] {$0.55$}; 
                          Alive ]
            \edge node[Action] {TEST};
            [.$\square$ 
                 \edge node[Pr] {$0.5$};
                 [.{Positive Test}
                    \edge node[Action] {WAIT};
                    [.$\square$  
                          \edge node[Pr] {$0.81$}; 
                          Dead
                          \edge node[Pr] {$0.19$}; 
                          Alive ]
                    \edge node[Action] {INJECT};
                    [.$\square$
                          \edge node[Pr] {$0.125$}; 
                          Dead
                          \edge node[Pr] {$0.875$}; 
                          Alive ]
                    ]
                 \edge node[Pr] {$0.5$};
                 [.{Negative Test}
                    \edge node[Action] {WAIT};
                    [.$\square$  
                          \edge node[Pr] {$0.09$}; 
                          Dead
                          \edge node[Pr] {$0.91$}; 
                          Alive ]
                    \edge node[Action] {INJECT};
                    [.$\square$
                          \edge node[Pr] {$0.725$}; 
                          Dead
                          \edge node[Pr] {$0.275$}; 
                          Alive ]
                    ]
             ]
            \edge node[Action] {INJECT};
            [.$\square$  
                          \edge node[Pr] {$0.425$}; 
                          Dead
                          \edge node[Pr] {$0.575$}; 
                          Alive ]
            ]
\end{tikzpicture}
\caption{A visualization of \EXTEND\ for a simple diagnostic scenario.  The initial belief state, $\varepsilon$, corresponds with the distribution assigning equal probability to the patient having the disease and not having the disease.  Edges corresponding with actions are labeled with the name of the action in a box.  For example, three actions are initially available: \emph{waiting} it out, running a diagnostic \emph{test}, and deciding to administer the \emph{injection}.  Square nodes ($\square$) represent the random belief state transitions that follow an action.  Edges leaving these nodes are labeled with the probabilities of the transitions. }
\label{decdiagram}
\end{figure}
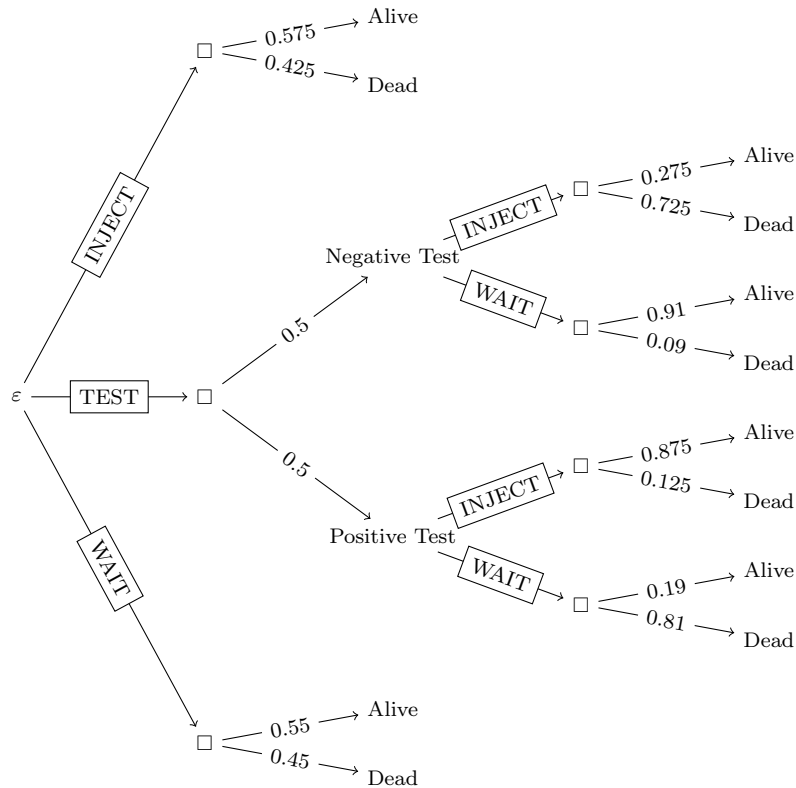

More concretely, we will assume that the patient is sick with the disease with probability $0.5$, and that, if the patient is sick, the injection will succeed in curing the patient with probability $0.95$, but will kill them with probability $0.8$ if they are not sick with the disease.  If the patient is sick, waiting things out is likely to succeed with probability $0.1$.  Finally, the diagnostic is accurate with probability $0.75$.  In Figure~\ref{decdiagram}, we present the corresponding belief state transition diagram capturing the behavior of \EXTEND.  

In order to understand the behavior of \POLICY\ at the initial belief state $\varepsilon$, we begin by studying its behavior after receiving test results.  In particular, having received a negative test result, the WAIT action succeeds with probability \linebreak $0.91$ and the INJECT action succeeds with probability $0.275$, and so \linebreak $\CAA(\text{Negative Test},\POLICY)$ chooses to WAIT with probability $\approx 0.77$.  (As $k \to \infty$, WAIT is asymptotically almost always chosen.)  On the other hand, having received a positive test result, $\CAA(\text{Positive Test},\POLICY)$ chooses to WAIT with probability $\approx 0.18$.  (Likewise, as $k\to \infty$, INJECT is asymptotically almost always chosen.)  These values imply that 
$
\SIM_{\varepsilon,\POLICY,\text{TEST}}$ accepts with probability $\approx 0.76$,
and so $\CAA(\varepsilon,\POLICY)$ assigns probability $0.40$, $0.29$, and $0.31$ to TEST, WAIT, and INJECT, respectively.  As $k \to \infty$, we see that we would asymptotically almost always choose to run a diagnostic test initially, as it significantly reduces our uncertainty.  This simple example only hints at the range of behaviors that can arise from \POLICY\ in complicated sequential decision tasks.
 
In the definition of \CAA, we choose a random action conditioned on future success.
Although we can find policies that optimize the probability of success  
by taking $k \to \infty$,
there are good reasons to not use a random action and instead use an initial policy that incorporates one's prior knowledge about the optimal action, much like it is advantageous to use a so-called admissible heuristic in search algorithms like $A^*$.
It is interesting to note that if the initial policy succeeds with very high probability then the \query\ expression above, viewed as algorithms, is even efficient.
Returning to a quote in Section~\ref{sec:converge}, Turing himself noted that knowing which actions to try first would 
``make the difference between a brilliant and a footling reasoner'' and that this knowledge might be 
``produced by the machine itself, \emph{e.g.}\ by scientific induction'' \cite[p.~458]{Tur50}.  
Indeed, recent work in Bayesian reinforcement learning has demonstrated the utility of more informative prior distributions on policies~\cite{DWRT10, WGRKT11}.

\subsection{Turing's insights}

There are a wide range of algorithms that have been developed to make decisions under uncertainty. 
Those familiar with economics and AI may see the connection between our analysis and the Markov Decision Process (MDP) formalism \cite{How60}, where the decision maker has full observability, 
and the Partially Observable MDP (POMDP) formalism, where the agent, like
above, has access to only part of the state (see \cite{Mon82} for a classic
survey and \cite{KLC98} for an early AI paper introducing the formalism).  
In AI, these subjects are studied in an area known as \emph{reinforcement learning}, which is in general the study of algorithms that can learn without receiving immediate feedback on their performance.  
(For a classic survey on reinforcement learning, see \cite{KLM96}.)  

A popular class of reinforcement learning algorithms are collectively called $Q$-learning, and were first introduced by Watkins \cite{Wat89}.  The simplest variants work by estimating the probability that an action $a$ leads eventually to success, starting from a belief state $b$ and assuming that all subsequent actions are chosen optimally.  This function, known as the $Q$- or action-value function is related to
\begin{align*}
\SIM_{b,\POLICY,a}
\end{align*}
when viewed as a function of belief state $b$ and action $a$.  In particular,
 the latter is an approximation to the former.
In $Q$-learning, estimates of this function are produced on the basis of experience interacting with the environment.
Under certain conditions, the estimate of the $Q$-function can be shown to converge to its true value~\cite{WD92}. 

It is instructive to compare the $Q$-learning algorithm to proposals that Turing himself made. In the course of a few pages in his 1950 article, Turing suggests mechanisms that learn by feedback in ways similar to methods in \emph{supervised learning} (immediate feedback) and reinforcement learning (delayed feedback):
\begin{quote}
We normally associate punishments and rewards with the teaching process. Some simple child machines can be constructed or programmed on this sort of principle. 
The machine has to be so constructed that events which shortly preceded the occurrence of a punishment signal are unlikely to be repeated, whereas a reward signal increased the probability of repetition of the events which led up to it.
\cite[p.~457]{Tur50}
\end{quote}
This quote describes behavior that, in broad
strokes, lines up well with how a reinforcement learning algorithm such as
$Q$-learning chooses which actions to take.
Yet such simple approaches to learning from trial and error, or teaching by reward and punishment, do not capture the most powerful ways humans learn and teach each other.  The last several decades of research in cognitive development \cite{Car09, Gop12, Sch12} have emphasized the many ways in which children's learning is more driven by intrinsic curiosity and motivation than by immediate external rewards and punishments, more like the discovery and refinement of scientific theories than the mechanisms of $Q$-learning or the process Turing describes above.
Indeed, a close analysis of the recursively defined \POLICY\ 
would show that its behavior is far more sophisticated than that of $Q$-learning.
In particular, $Q$-learning does not take advantage of the model \EXTEND, essentially assuming that there is no predictable structure in the evolution of the belief state.  On the other hand, \POLICY\ will perform information-gathering tasks, which could be construed as curiosity-driven, but are also rational acts that
may themselves improve the agent's future ability to act.

Turing himself also realized that more sophisticated learning mechanisms would be required, that a machine would have to learn from ``unemotional'' channels of communication, 
which, in the language of this section, 
would correspond with patterns in the observations themselves not directly linked to eventual success or failure.  
This type of \emph{unsupervised} learning would be useful if the goals or criteria for success changed, but the environment stayed the same. 
Turing envisioned that the memory store of a child-machine 
\begin{quote}
\quoteskip\ would be largely occupied with definitions and propositions.  The propositions would have various kinds of status, \emph{e.g.}\ well-established facts, conjectures, mathematically proved theorems, statements given by an authority, expressions having the logical form of proposition but not belief-value.  
\cite[p.~457]{Tur50}
\end{quote}
The idea of using a logical language as an underlying representation of knowledge has been studied since the early days of AI, and was even proposed as a means to achieve common-sense reasoning by contemporaries of Turing, such as McCarthy \cite{McC68}.
The problem of learning logical formulae from data, especially in domains with complex, discrete structure,
is actively pursued today by researchers in Statistical Relational Learning
\cite{GT07}
and Inductive Logic Programming \cite{Mug91}.

Turing imagined that the same collection of logical formulae would also pertain to decision-making:
\begin{quote}
Certain propositions may be described as `imperatives'.  
The machine should be so constructed that as soon as an imperative is classed as `well-established' the appropriate action automatically takes place.
\cite[p.~457]{Tur50}
\end{quote}
A similar mechanism would later be used in \emph{expert systems}, which first appeared in the 1960s and rose
to popularity in the 1980s as they demonstrated their usefulness and commercial viability.  (See \cite{LBFL93}
for a retrospective on one of the first successful expert system.)  

Having seen how \query\ can be used to make decisions under
uncertainty, we now conclude with some general thoughts about the use of
\query\ in common-sense reasoning.

\section{Towards common-sense reasoning}
\label{conclusion}

As we have seen, the \query\ framework can be used to model many common-sense reasoning tasks, and the underlying formalism owes much to Turing, as do several details of the approach. 
In many of the applications we have considered, the key step is providing \query\ with an appropriate
model---a generative description of the 
relevant aspects of nature.

In modeling, too, Turing was a pioneer. 
As evidenced by his diverse body of work across computation, statistics and even
morphogenesis \cite{Tur52},
Turing excelled in building simple models of complex natural phenomena.
In morphogenesis, in particular, 
his reaction-diffusion model proposed a particular sort of simplified chemical interactions as a way to understand visible
patterns on animals, but also potentially leaf arrangements and even aspects of embryo formation
\cite[p.~509]{Cop04}.
Turing hoped to make progress in understanding these biological phenomena by carefully analyzing 
simplified
mathematical models of
these natural systems.

The medical diagnosis model \DS\ that we examined in Section~\ref{sec:DS} 
is a crude attempt at the same sort of mechanical/computational
description of the natural patterns of co-occurrence of diseases and symptoms. As we have seen, using 
a generative model of these patterns as an input to \query, we can reason about unseen processes, like diseases,
from observed ones, like symptoms.  
We expect the inferences produced by \query\ to be diagnostically more useful
when
the generative model reflects 
a deep
scientific understanding of the  
mechanisms underlying the pattern of diseases and symptoms that we find in the human population.
But we also expect the inferences produced by \query\ to reflect natural patterns of common-sense reasoning among lay people when fed a model that, like \DS, represents a cruder state of uncertainty.

These explanations typify
\emph{computational theories}
in the
sense of Marr \cite{Mar82}, and especially the Bayesian accounts developed in Anderson's rational analyses \cite{And90},
Shepard's investigations of universal laws \cite{She87}, 
ideal observer models \linebreak \cite{Gei84, KY03},
and the work of Tenenbaum, Griffiths, and colleagues on
concept learning and generalization \cite{TG01}; see
also Oaksford and Chater's notion of Bayesian rationality \cite{OC98,OC07}.

A recent body of literature demonstrates that many human inferences and decisions in \emph{natural} situations are 
well predicted by probabilistic inference in models defined by simple generative descriptions of the 
underlying causal structure.
One set of examples concerns ``inverse planning'' in social cognition
\cite{BGT07, BST09, GBT09,UBM+09, Bak12} 
and in language \cite{GS12}. Using the approximate planning/decision-making framework discussed in
Section~\ref{sec:tom} as part of a generative model for human behavior, this research considers situations in which a human reasons about the goals of other agents having only observed their actions---hence the term \emph{inverse} planning---by assuming that the other agents are acting nearly optimally in attaining their goals.  
These approaches can lead to models that are good at making quantitative predictions of human
judgements about the intentions of others. 
As another example, the ``intuitive physics'' research by Hamrick and
Battaglia
\cite{HBT11, Ham12}
aims to explain human reasoning
about the physical world by positing that we use a crude ability to simulate simple physical models in our minds.
Other examples include 
pragmatics in language \cite{FG12, SG13}
and counterfactual reasoning \cite{GGLT12, MUSTT12}.
With all of these examples, 
there is a rather large gap between defining the given problem in that way and being able to computationally
solve it. But still there is substantial clarity brought about by the view of using \query\ along with a
generative description of underlying causal structure.

Of course, this raises the questions:  
how do we obtain such models? In particular, how can or should we build
them when they are not handed to us? And is there any hope of automating the process by which we, as scientists, invent such models upon mental reflection?
These are hard scientific problems, and we have addressed them in only a very narrow sense.
In Section~\ref{sec:latent}, we showed how the parameters to the \DS\ could be learned from data by
constructing a larger generative process, $\DS'$, wherein these parameters are also expressed as being uncertain.  We also showed, in Section~\ref{sec:definetti}, how the
conditional independence structure implicit in the \DS\ model could itself be learned, via inference in the model \RPD.

These examples suggest that one possible approach to learning models is via a more abstract version of the
sort of inference we have been describing, and this approach 
is roughly that taken
in ``theory-based Bayesian models'' approach of Griffiths, Kemp, and Tenenbaum \cite{GT06,
GKT08, KT08, GT09}.
Some examples include attempts to
learn structural forms \cite{TGK06}, and to learn a theory of causality \cite{GUT11}.
There are of course many other proposed approaches to learning models, some with flavors very different
from those considered in this paper.
Finally, the question of how to approach common-sense reasoning remains. Perhaps common-sense involves
knowing how to build one's own models, in a general enough setting to encompass all of experience. It is
clearly far too early to tell whether this, or any current approach, will succeed.

Although Turing did not frame his AI work in terms of conditioning, his generative models for morphogenesis did capture one of the key ideas presented here---that of
explaining a natural phenomenon via a detailed stochastic model of the underlying causal process.
More generally, given the wide range of Turing's ideas that appear together in the approach to AI we have
described, it is fascinating to speculate on what sort of synthesis Turing might have made, if he had had
the opportunity.

A tantalizing clue is offered by his wartime colleague I.~J.~Good: 
Near the end of his life, Turing was a member, along with several prominent neurologists, statisticians,
and physicists, of a small exclusive discussion group known as the \emph{Ratio Club}, 
named in part because of 
``the dependence of perception on the judging of ratios'' \cite[p.~101]{Goo91}.

One can only guess at how Turing might have combined his computational insight, statistical brilliance,
and passion for modeling natural phenomena into still further pursuits in AI.

\section*{Acknowledgements}

The authors would like to thank 
Nate Ackerman, 
Chris Baker,
Owain Evans,
Leslie Kaelbling,
Jonathan Malmaud,
Vikash Mansinghka, 
Timothy O'Donnell,
and Juliet Wagner
for very helpful discussions and critical feedback on drafts,
and Noah Goodman, Susan Holmes, Max Siegel, Andreas Stuhlm{\"u}ller, and Sandy Zabell for useful conversations. 
This publication was made possible through the support of grants from the John Templeton Foundation and Google. The opinions expressed in this publication are those of the authors and do not necessarily reflect the views of the John Templeton Foundation.  This paper was partially written while C.E.F.\ and D.M.R.\ were participants in the program \emph{Semantics and Syntax: A Legacy of Alan Turing} at the Isaac Newton Institute for the Mathematical Sciences.  D.M.R.\ is supported by a Newton International Fellowship and Emmanuel College.

\newpage
\begin{small}

\newcommand{\etalchar}[1]{$^{#1}$}
\def\cprime{$'$} \def\cprime{$'$}
  \def\polhk#1{\setbox0=\hbox{#1}{\ooalign{\hidewidth
  \lower1.5ex\hbox{`}\hidewidth\crcr\unhbox0}}}
  \def\polhk#1{\setbox0=\hbox{#1}{\ooalign{\hidewidth
  \lower1.5ex\hbox{`}\hidewidth\crcr\unhbox0}}} \def\cprime{$'$}
  \def\cprime{$'$} \def\cprime{$'$} \def\cprime{$'$}

\end{small}


\begin{thebibliography}{WGR{\etalchar{+}}11}

\bibitem[AB]{AB13}
{\sc J.~Avigad and V.~Brattka}.
\newblock Computability and analysis: The legacy of {A}lan {T}uring.
\newblock In R.~Downey, editor, {\em Turing's Legacy}. Cambridge University
  Press, Cambridge, UK.

\bibitem[AFR11]{AFR11}
{\sc N.~L. Ackerman, C.~E. Freer, and D.~M. Roy}.
\newblock Noncomputable conditional distributions.
\newblock In {\em {Proceedings of the 26th Annual IEEE Symposium on Logic in
  Computer Science (LICS 2011)}}, pages 107--116. IEEE Computer Society, 2011.

\bibitem[And90]{And90}
{\sc J.~R. Anderson}.
\newblock {\em {The Adaptive Character of Thought}}.
\newblock Erlbaum, Hillsdale, NJ, 1990.

\bibitem[Bak12]{Bak12}
{\sc C.~L. Baker}.
\newblock {\em {Bayesian Theory of Mind: Modeling Human Reasoning about
  Beliefs, Desires, Goals, and Social Relations}}.
\newblock PhD thesis, Massachusetts Institute of Technology, 2012.

\bibitem[Bar98]{Bar98}
{\sc A.~R. Barron}.
\newblock Information-theoretic characterization of {B}ayes performance and the
  choice of priors in parametric and nonparametric problems.
\newblock In J.~M. Bernardo, J.~O. Berger, A.~P. Dawid, and A.~F.~M. Smith,
  editors, {\em {Bayesian Statistics 6: Proceedings of the Sixth Valencia
  International Meeting}}, pages 27--52, 1998.

\bibitem[Bel57]{Bel57}
{\sc R.~Bellman}.
\newblock {\em {Dynamic Programming}}.
\newblock Princeton University Press, Princeton, NJ, 1957.

\bibitem[BGT07]{BGT07}
{\sc C.~L. Baker, N.~D. Goodman, and J.~B. Tenenbaum}.
\newblock Theory-based social goal inference.
\newblock In {\em {Proceedings of the 30th Annual Conference of the Cognitive
  Science Society}}, pages 1447--1452, 2007.

\bibitem[BJ03]{BJ03}
{\sc F.~R. Bach and M.~I. Jordan}.
\newblock Learning graphical models with {M}ercer kernels.
\newblock In S.~Becker, S.~Thrun, and K.~Obermayer, editors, {\em {Advances in
  Neural Information Processing Systems 15 (NIPS 2002)}}, pages 1009--1016. The
  MIT Press, Cambridge, MA, 2003.

\bibitem[Bla97]{Bla97}
{\sc J.~Blanck}.
\newblock Domain representability of metric spaces.
\newblock {\em Annals of Pure and Applied Logic}, 83(3):225 -- 247, 1997.

\bibitem[BST09]{BST09}
{\sc C.~L. Baker, R.~Saxe, and J.~B. Tenenbaum}.
\newblock Action understanding as inverse planning.
\newblock {\em Cognition}, 113(3):329--349, 2009.

\bibitem[Cam11]{Cam11}
{\sc C.~F. Camerer}.
\newblock {\em {Behavioral Game Theory: Experiments in Strategic Interaction}}.
\newblock The Roundtable Series in Behavioral Economics. Princeton University
  Press, 2011.

\bibitem[Car09]{Car09}
{\sc S.~Carey}.
\newblock {\em {The Origin of Concepts}}.
\newblock Oxford University Press, New York, 2009.

\bibitem[Coo90]{Coo90}
{\sc G.~F. Cooper}.
\newblock The computational complexity of probabilistic inference using
  {B}ayesian belief networks.
\newblock {\em Artificial Intelligence}, 42(2-3):393--405, 1990.

\bibitem[Cop04]{Cop04}
B.~J. Copeland, editor.
\newblock {\em {The Essential {T}uring: Seminal Writings in Computing, Logic,
  Philosophy, Artificial Intelligence, and Artificial Life: Plus the Secrets of
  Enigma}}.
\newblock Oxford University Press, Oxford, 2004.

\bibitem[CP96]{CP96}
{\sc B.~J. Copeland and D.~Proudfoot}.
\newblock On {A}lan {T}uring's anticipation of connectionism.
\newblock {\em Synthese}, 108(3):pp. 361--377, 1996.

\bibitem[CSH08]{CSH08}
{\sc V.~Chandrasekaran, N.~Srebro, and P.~Harsha}.
\newblock Complexity of inference in graphical models.
\newblock In {\em {Proceedings of the Twenty Fourth Conference on Uncertainty
  in Artificial Intelligence (UAI 2008)}}, pages 70--78, Corvalis, Oregon,
  2008. AUAI Press.

\bibitem[DeG05]{DeG05}
{\sc M.~H. DeGroot}.
\newblock {\em {Optimal Statistical Decisions}}.
\newblock Wiley Classics Library. Wiley, 2005.

\bibitem[DKLR00]{DKLR00}
{\sc P.~Dagum, R.~Karp, M.~Luby, and S.~Ross}.
\newblock An optimal algorithm for {M}onte {C}arlo estimation.
\newblock {\em SIAM Journal on Computing}, 29(5):1484--1496, 2000.

\bibitem[DL93]{DL93}
{\sc P.~Dagum and M.~Luby}.
\newblock Approximating probabilistic inference in {B}ayesian belief networks
  is {NP}-hard.
\newblock {\em Artificial Intelligence}, 60(1):141--153, 1993.

\bibitem[dMSS56]{dMSS56}
{\sc K.~d{{e Leeuw}}, E.~F. Moore, C.~E. Shannon, and N.~Shapiro}.
\newblock Computability by probabilistic machines.
\newblock In {\em Automata {S}tudies}, Annals of {M}athematical {S}tudies, no.
  34, pages 183--212. Princeton University Press, Princeton, N. J., 1956.

\bibitem[DWRT10]{DWRT10}
{\sc F.~{Doshi-Velez}, D.~Wingate, N.~Roy, and J.~Tenenbaum}.
\newblock Nonparametric {B}ayesian policy priors for reinforcement learning.
\newblock In J.~Lafferty, C.~K.~I. Williams, J.~Shawe-Taylor, R.~S. Zemel, and
  A.~Culotta, editors, {\em {Advances in Neural Information Processing Systems
  23 (NIPS 2010)}}, pages 532--540. 2010.

\bibitem[Eda96]{Eda96}
{\sc A.~Edalat}.
\newblock The {S}cott topology induces the weak topology.
\newblock In {\em {11th Annual {IEEE} {S}ymposium on {L}ogic in {C}omputer
  {S}cience (LICS 1996)}}, pages 372--381. IEEE Computer Society Press, Los
  Alamitos, CA, 1996.

\bibitem[EH98]{EH98}
{\sc A.~Edalat and R.~Heckmann}.
\newblock A computational model for metric spaces.
\newblock {\em {Theoretical Computer Science}}, 193(1-2):53--73, 1998.

\bibitem[FG12]{FG12}
{\sc M.~C. Frank and N.~D. Goodman}.
\newblock Predicting pragmatic reasoning in language games.
\newblock {\em Science}, 336(6084):998, 2012.

\bibitem[G{\'a}c05]{Gac05}
{\sc P.~G{\'a}cs}.
\newblock Uniform test of algorithmic randomness over a general space.
\newblock {\em Theoretical Computer Science}, 341(1-3):91--137, 2005.

\bibitem[GBT09]{GBT09}
{\sc N.~D. Goodman, C.~L. Baker, and J.~B. Tenenbaum}.
\newblock Cause and intent: Social reasoning in causal learning.
\newblock In {\em {Proceedings of the 31st Annual Conference of the Cognitive
  Science Society}}, pages 2759--2764, 2009.

\bibitem[Gei84]{Gei84}
{\sc W.~S. Geisler}.
\newblock Physical limits of acuity and hyperacuity.
\newblock {\em Journal of the Optical Society of America A}, 1(7):775--782,
  1984.

\bibitem[GG02]{GG02}
{\sc D.~G. Goldstein and G.~Gigerenzer}.
\newblock Models of ecological rationality: The recognition heuristic.
\newblock {\em Psychological Review}, 109(1):75--90, January 2002.

\bibitem[GG12]{GG12}
{\sc T.~Gerstenberg and N.~D. Goodman}.
\newblock Ping pong in {C}hurch: Productive use of concepts in human
  probabilistic inference.
\newblock In N.~Miyake, D.~Peebles, and R.~P. Cooper, editors, {\em
  {Proceedings of the Thirty-Fourth {A}nnual {C}onference of the {C}ognitive
  {S}cience {S}ociety}}. {A}ustin, {TX}: Cognitive {S}cience {S}ociety, 2012.

\bibitem[GGLT12]{GGLT12}
{\sc T.~Gerstenberg, N.~D. Goodman, D.~A. Lagnado, and J.~B. Tenenbaum}.
\newblock Noisy {N}ewtons: {U}nifying process and dependency accounts of causal
  attribution.
\newblock In N.~Miyake, D.~Peebles, and R.~P. Cooper, editors, {\em
  {Proceedings of the Thirty-Fourth Annual Conference of the Cognitive Science
  Society}}. {A}ustin, {TX}: Cognitive {S}cience {S}ociety, 2012.

\bibitem[GHR10]{GHR10}
{\sc S.~Galatolo, M.~Hoyrup, and C.~Rojas}.
\newblock Effective symbolic dynamics, random points, statistical behavior,
  complexity and entropy.
\newblock {\em Information and Computation}, 208(1):23--41, 2010.

\bibitem[GKT08]{GKT08}
{\sc T.~L. Griffiths, C.~Kemp, and J.~B. Tenenbaum}.
\newblock Bayesian models of cognition.
\newblock In {\em {Cambridge Handbook of Computational Cognitive Modeling}}.
  Cambridge University Press, 2008.

\bibitem[GMR{\etalchar{+}}08]{GMRBT08}
{\sc N.~D. Goodman, V.~K. Mansinghka, D.~M. Roy, K.~Bonawitz, and J.~B.
  Tenenbaum}.
\newblock Church: A language for generative models.
\newblock In {\em {Proceedings of the Twenty-Fourth Conference on Uncertainty
  in Artificial Intelligence (UAI 2008)}}, pages 220--229, Corvalis, Oregon,
  2008. AUAI Press.

\bibitem[Goo61]{Goo61}
{\sc I.~J. Good}.
\newblock A causal calculus. {I}.
\newblock {\em The British Journal for the Philosophy of Science}, 11:305--318,
  1961.

\bibitem[Goo68]{Goo68}
{\sc I.~J. Good}.
\newblock Corroboration, explanation, evolving probability, simplicity and a
  sharpened razor.
\newblock {\em The British Journal for the Philosophy of Science},
  19(2):123--143, 1968.

\bibitem[Goo75]{Goo75}
{\sc I.~J. Good}.
\newblock Explicativity, corroboration, and the relative odds of hypotheses.
\newblock {\em Synthese}, 30(1):39--73, 1975.

\bibitem[Goo79]{Goo79}
{\sc I.~J. Good}.
\newblock {A}. {M}. {T}uring's statistical work in {W}orld {W}ar {II}.
\newblock {\em Biometrika}, 66(2):393--396, 1979.

\bibitem[Goo91]{Goo91}
{\sc I.~J. Good}.
\newblock Weight of evidence and the {B}ayesian likelihood ratio.
\newblock In C.~G.~G. Aitken and D.~A. Stoney, editors, {\em {The Use Of
  Statistics In Forensic Science}}. Ellis Horwood, Chichester, 1991.

\bibitem[Goo00]{Goo00}
{\sc I.~J. Good}.
\newblock Turing's anticipation of empirical {B}ayes in connection with the
  cryptanalysis of the naval {E}nigma.
\newblock {\em Journal of Statistical Computation and Simulation},
  66(2):101--111, 2000.

\bibitem[Gop12]{Gop12}
{\sc A.~Gopnik}.
\newblock Scientific thinking in young children: Theoretical advances,
  empirical research, and policy implications.
\newblock {\em Science}, 337(6102):1623--1627, 2012.

\bibitem[GS12]{GS12}
{\sc N.~D. Goodman and A.~Stuhlm{\"u}ller}.
\newblock Knowledge and implicature: Modeling language understanding as social
  cognition.
\newblock In N.~Miyake, D.~Peebles, and R.~P. Cooper, editors, {\em
  {Proceedings of the Thirty-Fourth {A}nnual {C}onference of the {C}ognitive
  {S}cience {S}ociety}}. {A}ustin, {TX}: Cognitive {S}cience {S}ociety, 2012.

\bibitem[GSW07]{GSW07}
{\sc T.~Grubba, M.~Schr{\"o}der, and K.~Weihrauch}.
\newblock Computable metrization.
\newblock {\em Mathematical Logic Quarterly}, 53(4-5):381--395, 2007.

\bibitem[GT05]{GT05}
{\sc T.~L. Griffiths and J.~B. Tenenbaum}.
\newblock Structure and strength in causal induction.
\newblock {\em Cognitive Psychology}, 51(4):334--384, 2005.

\bibitem[GT06]{GT06}
{\sc T.~L. Griffiths and J.~B. Tenenbaum}.
\newblock Optimal predictions in everyday cognition.
\newblock {\em Psychological Science}, 17(9):767--773, 2006.

\bibitem[GT07]{GT07}
{\sc L.~Getoor and B.~Taskar}.
\newblock {\em {Introduction to Statistical Relational Learning}}.
\newblock The MIT Press, 2007.

\bibitem[GT09]{GT09}
{\sc T.~L. Griffiths and J.~B. Tenenbaum}.
\newblock Theory-based causal induction.
\newblock {\em Psychological Review}, 116(4):661--716, 2009.

\bibitem[GT12]{GT12}
{\sc N.~D. Goodman and J.~B. Tenenbaum}.
\newblock The probabilistic language of thought.
\newblock In preparation, 2012.

\bibitem[GTFG08]{GTFG08}
{\sc N.~D. Goodman, J.~B. Tenenbaum, J.~Feldman, and T.~L. Griffiths}.
\newblock A rational analysis of rule-based concept learning.
\newblock {\em Cognitive Science}, 32(1):108--154, 2008.

\bibitem[GTO11]{GTO11}
{\sc N.~D. Goodman, J.~B. Tenenbaum, and T.~J. O'Donnell}.
\newblock Probabilistic models of cognition.
\newblock {\em Church wiki}, 2011.
\newblock
  \url{http://projects.csail.mit.edu/church/wiki/Probabilistic_Models_of_Cogni%
tion}.

\bibitem[GUT11]{GUT11}
{\sc N.~D. Goodman, T.~D. Ullman, and J.~B. Tenenbaum}.
\newblock Learning a theory of causality.
\newblock {\em Psychological Review}, 118(1):110--119, 2011.

\bibitem[Ham12]{Ham12}
{\sc J.~Hamrick}.
\newblock {Physical Reasoning in Complex Scenes is Sensitive to Mass}.
\newblock Master of Engineering thesis, Massachusetts Institute of Technology,
  Cambridge, MA, 2012.

\bibitem[HBT11]{HBT11}
{\sc J.~Hamrick, P.~W. Battaglia, and J.~B. Tenenbaum}.
\newblock Internal physics models guide probabilistic judgments about object
  dynamics.
\newblock In C.~Carlson, C.~H{\"o}lscher, and T.~Shipley, editors, {\em
  {Proceedings of the Thirty-Third Annual Conference of the Cognitive Science
  Society}}, pages 1545--1550. {A}ustin, {TX}: Cognitive {S}cience {S}ociety,
  2011.

\bibitem[Hem02]{Hem02}
{\sc A.~Hemmerling}.
\newblock Effective metric spaces and representations of the reals.
\newblock {\em Theoretical Computer Science}, 284(2):347--372, 2002.

\bibitem[Hod97]{Hod97}
{\sc A.~Hodges}.
\newblock {\em {Turing: A Natural Philosopher}}.
\newblock Phoenix, London, 1997.

\bibitem[How60]{How60}
{\sc R.~A. Howard}.
\newblock {\em {Dynamic Programming and {M}arkov Processes}}.
\newblock The MIT Press, Cambridge, MA, 1960.

\bibitem[Kal02]{Kal02}
{\sc O.~Kallenberg}.
\newblock {\em Foundations of modern probability}.
\newblock Probability and its Applications. Springer, New York, 2nd edition,
  2002.

\bibitem[KGT08]{KGT08}
{\sc C.~Kemp, N.~D. Goodman, and J.~B. Tenenbaum}.
\newblock Learning and using relational theories.
\newblock In {\em {Advances in Neural Information Processing Systems 20 (NIPS
  2007) }}, 2008.

\bibitem[KLC98]{KLC98}
{\sc L.~P. Kaelbling, M.~L. Littman, and A.~R. Cassandra}.
\newblock Planning and acting in partially observable stochastic domains.
\newblock {\em Artificial Intelligence}, 101:99--134, 1998.

\bibitem[KLM96]{KLM96}
{\sc L.~P. Kaelbling, M.~L. Littman, and A.~W. Moore}.
\newblock Reinforcement learning: A survey.
\newblock {\em Journal of Artificial Intelligence Research}, 4:237--285, 1996.

\bibitem[KSBT07]{KSBT07}
{\sc C.~Kemp, P.~Shafto, A.~Berke, and J.~B. Tenenbaum}.
\newblock Combining causal and similarity-based reasoning.
\newblock In B.~Sch\"{o}lkopf, J.~Platt, and T.~Hoffman, editors, {\em
  {Advances in Neural Information Processing Systems 19 (NIPS 2006)}}, pages
  681--688. The MIT Press, Cambridge, MA, 2007.

\bibitem[KT08]{KT08}
{\sc C.~Kemp and J.~B. Tenenbaum}.
\newblock The discovery of structural form.
\newblock {\em Proceedings of the National Academy of Sciences},
  105(31):10687--10692, 2008.

\bibitem[KY03]{KY03}
{\sc D.~Kersten and A.~Yuille}.
\newblock Bayesian models of object perception.
\newblock {\em Current Opinion in Neurobiology}, 13(2):150--158, 2003.

\bibitem[LBFL93]{LBFL93}
{\sc R.~K. Lindsay, B.~G. Buchanan, E.~A. Feigenbaum, and J.~Lederberg}.
\newblock {DENDRAL}: {A} case study of the first expert system for scientific
  hypothesis formation.
\newblock {\em Artificial Intelligence}, 61(2):209--261, 1993.

\bibitem[Luc59]{Luc59}
{\sc R.~D. Luce}.
\newblock {\em {Individual Choice Behavior}}.
\newblock John Wiley, New York, 1959.

\bibitem[Luc77]{Luc77}
{\sc R.~D. Luce}.
\newblock The choice axiom after twenty years.
\newblock {\em Journal of Mathematical Psychology}, 15(3):215--233, 1977.

\bibitem[Mac03]{Mac03}
{\sc D.~J.~C. MacKay}.
\newblock {\em {Information Theory, Inference, and Learning Algorithms}}.
\newblock Cambridge University Press, Cambridge, UK, 2003.

\bibitem[Man09]{Man09}
{\sc V.~K. Mansinghka}.
\newblock {\em {Natively Probabilistic Computation}}.
\newblock PhD thesis, Massachusetts Institute of Technology, 2009.

\bibitem[Man11]{Man11}
{\sc V.~K. Mansinghka}.
\newblock Beyond calculation: Probabilistic computing machines and universal
  stochastic inference.
\newblock {\em NIPS Philosophy and Machine Learning Workshop}, 2011.

\bibitem[Mar82]{Mar82}
{\sc D.~Marr}.
\newblock {\em Vision}.
\newblock Freeman, San Francisco, 1982.

\bibitem[McC68]{McC68}
{\sc J.~McCarthy}.
\newblock Programs with common sense.
\newblock In {\em {Semantic Information Processing}}, pages 403--418. The MIT
  Press, 1968.

\bibitem[MHC03]{MHC03}
{\sc O.~Madani, S.~Hanks, and A.~Condon}.
\newblock On the undecidability of probabilistic planning and related
  stochastic optimization problems.
\newblock {\em Artificial Intelligence}, 147(1--2):5--34, 2003.

\bibitem[MJT08]{MJT08}
{\sc V.~K. Mansinghka, E.~Jonas, and J.~B. Tenenbaum}.
\newblock Stochastic digital circuits for probabilistic inference.
\newblock Technical Report MIT-CSAIL-TR-2008-069, Massachusetts Institute of
  Technology, 2008.

\bibitem[MKTG06]{MKTG06}
{\sc V.~K. Mansinghka, C.~Kemp, J.~B. Tenenbaum, and T.~L. Griffiths}.
\newblock Structured priors for structure learning.
\newblock In {\em {Proceedings of the Twenty-Second Conference on Uncertainty
  in Artificial Intelligence (UAI 2006)}}, pages 324--331, Arlington, Virginia,
  2006. AUAI Press.

\bibitem[Mon82]{Mon82}
{\sc G.~E. Monahan}.
\newblock A survey of {P}artially {O}bservable {M}arkov {D}ecision {P}rocesses:
  Theory, models, and algorithms.
\newblock {\em Management Science}, 28(1):pp. 1--16, 1982.

\bibitem[MP43]{MP43}
{\sc W.~S. McCulloch and W.~Pitts}.
\newblock A logical calculus of the ideas immanent in nervous activity.
\newblock {\em Bulletin of Mathematical Biology}, 5(4):115--133, 1943.

\bibitem[MR]{MR13}
{\sc V.~K. Mansinghka and D.~M. Roy}.
\newblock Stochastic inference machines.
\newblock In preparation.

\bibitem[Mug91]{Mug91}
{\sc S.~Muggleton}.
\newblock Inductive logic programming.
\newblock {\em New Generation Computing}, 8(4):295--318, 1991.

\bibitem[MUS{\etalchar{+}}12]{MUSTT12}
{\sc J.~McCoy, T.~D. Ullman, A.~Stuhlm{\"u}ller, T.~Gerstenberg, and J.~B.
  Tenenbaum}.
\newblock Why blame {B}ob? {P}robabilistic generative models, counterfactual
  reasoning, and blame attribution.
\newblock In N.~Miyake, D.~Peebles, and R.~P. Cooper, editors, {\em
  {Proceedings of the Thirty-Fourth Annual Conference of the Cognitive Science
  Society}}. {A}ustin, {TX}: Cognitive {S}cience {S}ociety, 2012.

\bibitem[OC98]{OC98}
M.~Oaksford and N.~Chater, editors.
\newblock {\em {Rational Models of Cognition}}.
\newblock Oxford University Press, Oxford, 1998.

\bibitem[OC07]{OC07}
{\sc M.~Oaksford and N.~Chater}.
\newblock {\em {Bayesian Rationality: The Probabilistic Approach to Human
  Reasoning}}.
\newblock Oxford University Press, New York, 2007.

\bibitem[Pea88]{Pea88}
{\sc J.~Pearl}.
\newblock {\em {Probabilistic Reasoning in Intelligent Systems: {N}etworks of
  Plausible Inference}}.
\newblock Morgan Kaufmann, San Francisco, 1988.

\bibitem[Pea04]{Pea04}
{\sc J.~Pearl}.
\newblock Graphical models for probabilistic and causal reasoning.
\newblock In A.~B. Tucker, editor, {\em {Computer Science Handbook}}. CRC
  Press, 2nd edition, 2004.

\bibitem[Pfa79]{Pfa79}
{\sc J.~Pfanzagl}.
\newblock Conditional distributions as derivatives.
\newblock {\em The Annals of Probability}, 7(6):1046--1050, 1979.

\bibitem[PT87]{PT87}
{\sc C.~H. Papadimitriou and J.~N. Tsitsiklis}.
\newblock The complexity of {M}arkov {D}ecision {P}rocesses.
\newblock {\em Mathematics of Operations Research}, 12(3):441--450, 1987.

\bibitem[Rao88]{Rao88}
{\sc M.~M. Rao}.
\newblock Paradoxes in conditional probability.
\newblock {\em Journal of Multivariate Analysis}, 27(2):434--446, 1988.

\bibitem[Rao05]{Rao05}
{\sc M.~M. Rao}.
\newblock {\em Conditional measures and applications}, volume 271 of {\em Pure
  and Applied Mathematics}.
\newblock Chapman \& Hall/CRC, Boca Raton, FL, 2nd edition, 2005.

\bibitem[RH11]{RH11}
{\sc S.~Rathmanner and M.~Hutter}.
\newblock A philosophical treatise of universal induction.
\newblock {\em Entropy}, 13(6):1076--1136, 2011.

\bibitem[Roy11]{Roy11}
{\sc D.~M. Roy}.
\newblock {\em {Computability, Inference and Modeling in Probabilistic
  Programming}}.
\newblock PhD thesis, Massachusetts Institute of Technology, 2011.

\bibitem[Sch07]{Sch07}
{\sc M.~Schr{\"o}der}.
\newblock Admissible representations for probability measures.
\newblock {\em Mathematical Logic Quarterly}, 53(4-5):431--445, 2007.

\bibitem[Sch12]{Sch12}
{\sc L.~Schulz}.
\newblock The origins of inquiry: Inductive inference and exploration in early
  childhood.
\newblock {\em Trends in Cognitive Sciences}, 16(7):382--389, 2012.

\bibitem[SG]{SG13}
{\sc A.~Stuhlm{\"u}ller and N.~D. Goodman}.
\newblock Reasoning about reasoning by nested conditioning: Modeling theory of
  mind with probabilistic programs.
\newblock Submitted.

\bibitem[SG92]{SG92}
{\sc A.~F.~M. Smith and A.~E. Gelfand}.
\newblock Bayesian statistics without tears: A sampling-resampling perspective.
\newblock {\em The American Statistician}, 46(2):pp. 84--88, 1992.

\bibitem[SG12]{SG12}
{\sc A.~Stuhlm{\"u}ller and N.~D. Goodman}.
\newblock A dynamic programming algorithm for inference in recursive
  probabilistic programs.
\newblock {\em Second Statistical Relational AI workshop at UAI 2012
  (StaRAI-12)}, 2012.

\bibitem[She87]{She87}
{\sc R.~N. Shepard}.
\newblock Toward a universal law of generalization for psychological science.
\newblock {\em Science}, 237(4820):1317--1323, 1987.

\bibitem[SMH{\etalchar{+}}91]{SMH+91}
{\sc M.~A. Shwe, B.~Middleton, D.~E. Heckerman, M.~Henrion, E.~J. Horvitz,
  H.~P. Lehmann, and G.~F. Cooper}.
\newblock Probabilistic diagnosis using a reformulation of the
  {INTERNIST-1/QMR} knowledge base.
\newblock {\em Methods of Information in Medicine}, 30:241--255, 1991.

\bibitem[Sol64]{Sol64}
{\sc R.~J. Solomonoff}.
\newblock A formal theory of inductive inference: Parts {I} and {II}.
\newblock {\em Information and Control}, 7(1):1--22 and 224--254, 1964.

\bibitem[Teu02]{Teu02}
{\sc C.~Teuscher}.
\newblock {\em {Turing's Connectionism: An Investigation of Neural Network
  Architectures}}.
\newblock Springer-Verlag, London, 2002.

\bibitem[TG01]{TG01}
{\sc J.~B. Tenenbaum and T.~L. Griffiths}.
\newblock Generalization, similarity, and {B}ayesian inference.
\newblock {\em Behavioral and Brain Sciences}, 24(4):629--640, 2001.

\bibitem[TGK06]{TGK06}
{\sc J.~B. Tenenbaum, T.~L. Griffiths, and C.~Kemp}.
\newblock Theory-based {B}ayesian models of inductive learning and reasoning.
\newblock {\em Trends in Cognitive Sciences}, 10:309--318, 2006.

\bibitem[THS06]{THS06}
{\sc M.~Toussaint, S.~Harmeling, and A.~Storkey}.
\newblock Probabilistic inference for solving {(PO)MDPs}.
\newblock Technical Report EDI-INF-RR-0934, University of Edinburgh, School of
  Informatics, 2006.

\bibitem[Tju74]{Tju74}
{\sc T.~Tjur}.
\newblock {\em {Conditional Probability Distributions}}.
\newblock Lecture Notes, no. 2. Institute of Mathematical Statistics,
  University of Copenhagen, Copenhagen, 1974.

\bibitem[Tju75]{Tju75}
{\sc T.~Tjur}.
\newblock {\em {A Constructive Definition of Conditional Distributions}}.
\newblock Preprint 13. Institute of Mathematical Statistics, University of
  Copenhagen, Copenhagen, 1975.

\bibitem[Tju80]{Tju80}
{\sc T.~Tjur}.
\newblock {\em {Probability Based on {R}adon Measures}}.
\newblock Wiley Series in Probability and Mathematical Statistics. John Wiley
  \& Sons Ltd., Chichester, 1980.

\bibitem[TKGG11]{TKGG11}
{\sc J.~B. Tenenbaum, C.~Kemp, T.~L. Griffiths, and N.~D. Goodman}.
\newblock How to grow a mind: Statistics, structure, and abstraction.
\newblock {\em Science}, 331(6022):1279--1285, 2011.

\bibitem[Tur36]{Tur36}
{\sc A.~M. Turing}.
\newblock On computable numbers, with an application to the
  {E}ntscheidungsproblem.
\newblock {\em Proceedings of the London Mathematical Society}, 42(1):230--265,
  1936.

\bibitem[Tur39]{Tur39}
{\sc A.~M. Turing}.
\newblock Systems of logic based on ordinals.
\newblock {\em Proceedings of the London Mathematical Society}, 45(1):161--228,
  1939.

\bibitem[Tur48]{Tur48}
{\sc A.~M. Turing}.
\newblock {\em {Intelligent Machinery}}.
\newblock National Physical Laboratory Report. 1948.

\bibitem[Tur50]{Tur50}
{\sc A.~M. Turing}.
\newblock Computing machinery and intelligence.
\newblock {\em Mind}, 59:433--460, 1950.

\bibitem[Tur52]{Tur52}
{\sc A.~M. Turing}.
\newblock The chemical basis of morphogenesis.
\newblock {\em Philosophical Transactions of the Royal Society of London.
  Series B, Biological Sciences}, 237(641):37--72, 1952.

\bibitem[Tur96]{Tur96}
{\sc A.~M. Turing}.
\newblock Intelligent machinery, a heretical theory.
\newblock {\em Philosophia Mathematica. Philosophy of Mathematics, its
  Learning, and its Applications. Series III}, 4(3):256--260, 1996.
\newblock Originally a radio presentation, 1951.

\bibitem[Tur12]{Tur12}
{\sc A.~M. Turing}.
\newblock {\em {The Applications of Probability to Cryptography, c. 1941}}.
\newblock UK National Archives, HW 25/37. 2012.

\bibitem[UBM{\etalchar{+}}09]{UBM+09}
{\sc T.~D. Ullman, C.~L. Baker, O.~Macindoe, O.~Evans, N.~D. Goodman, and J.~B.
  Tenenbaum}.
\newblock Help or hinder: {B}ayesian models of social goal inference.
\newblock In {\em {Advances in Neural Information Processing Systems 22 (NIPS
  2009)}}, pages 1874--1882, 2009.

\bibitem[Wat89]{Wat89}
{\sc C.~J. C.~H. Watkins}.
\newblock {\em {Learning from Delayed Rewards}}.
\newblock PhD thesis, King's College, University of Cambridge, 1989.

\bibitem[WD92]{WD92}
{\sc C.~J. C.~H. Watkins and P.~Dayan}.
\newblock {\emph{Q}}-{L}earning.
\newblock {\em Machine Learning}, 8:279--292, 1992.

\bibitem[Wei93]{Wei93}
{\sc K.~Weihrauch}.
\newblock Computability on computable metric spaces.
\newblock {\em Theoretical Computer Science}, 113(2):191--210, 1993.

\bibitem[Wei99]{Wei99}
{\sc K.~Weihrauch}.
\newblock Computability on the probability measures on the {B}orel sets of the
  unit interval.
\newblock {\em Theoretical Computer Science}, 219(1-2):421--437, 1999.

\bibitem[Wei00]{Wei00}
{\sc K.~Weihrauch}.
\newblock {\em {Computable Analysis: An Introduction}}.
\newblock Texts in Theoretical Computer Science, An EATCS Series.
  Springer-Verlag, Berlin, 2000.

\bibitem[WGR{\etalchar{+}}11]{WGRKT11}
{\sc D.~Wingate, N.~D. Goodman, D.~M. Roy, L.~P. Kaelbling, and J.~B.
  Tenenbaum}.
\newblock Bayesian policy search with policy priors.
\newblock In T.~Walsh, editor, {\em {Proceedings of the Twenty-Second
  International Joint Conference on Artificial Intelligence (IJCAI)}}, Menlo
  Park, CA, 2011. AAAI Press.

\bibitem[WGSS11]{WGSS11}
{\sc D.~Wingate, N.~D. Goodman, A.~Stuhlm{\"u}ller, and J.~M. Siskind}.
\newblock Nonstandard interpretations of probabilistic programs for efficient
  inference.
\newblock In {\em {Advances in Neural Information Processing Systems 24 (NIPS
  2011) }}, 2011.

\bibitem[WSG11]{WSG11}
{\sc D.~Wingate, A.~Stuhlm{\"u}ller, and N.~D. Goodman}.
\newblock Lightweight implementations of probabilistic programming languages
  via transformational compilation.
\newblock In {\em {Proceedings of the Fourteenth International Conference on
  Artificial Intelligence and Statistics (AISTATS)}}, volume~15 of {\em Journal
  of Machine Learning Research: Workshop and Conference Proceedings}, pages
  770--778, 2011.

\bibitem[Yam99]{Yam99}
{\sc T.~Yamakami}.
\newblock Polynomial time samplable distributions.
\newblock {\em Journal of Complexity}, 15(4):557--574, 1999.

\bibitem[Zab95]{Zab95}
{\sc S.~L. Zabell}.
\newblock Alan {T}uring and the central limit theorem.
\newblock {\em American Mathematical Monthly}, 102(6):483--494, 1995.

\bibitem[Zab12]{Zab12}
{\sc S.~L. Zabell}.
\newblock Commentary on {A}lan {M}. {T}uring: The applications of probability
  to cryptography.
\newblock {\em Cryptologia}, 36(3):191--214, 2012.

\end{thebibliography}
\end{document}